\PassOptionsToPackage{table,xcdraw}{xcolor}
\documentclass[10pt,twocolumn,letterpaper]{article}

\usepackage{cvpr}              
\usepackage{booktabs}   
\usepackage{multirow}   
\usepackage{graphicx}   
\usepackage{array}      

\usepackage{bm}
\usepackage{makecell} 
\usepackage{amsmath}
\usepackage{wasysym}
\usepackage{pifont}
\usepackage{amssymb}

\newcommand{\xmark}{\textcolor{red!70!black}{\ding{55}}}   
\newcommand{\cmark}{\textcolor{green!60!black}{\ding{51}}} 
\usepackage{bm}
\usepackage{fancyhdr}

\definecolor{cvprblue}{rgb}{0.21,0.49,0.74}
\usepackage[pagebackref,breaklinks,colorlinks,allcolors=cvprblue]{hyperref}
\usepackage{tikz}

\def\paperID{*} 
\def\confName{CVPR}
\def\confYear{2026}

\title{\texttt{UniVBench}: Towards Unified Evaluation for Video Foundation Models}
\author{
  Jianhui Wei$^{1,2}$~\footnotemark[1] ~\footnotemark[4],\quad  
  Xiaotian Zhang$^{1,2}$~\footnotemark[1] ~\footnotemark[4],\quad 
  Yichen Li$^{1,2}$~\footnotemark[1] ~\footnotemark[4],\quad 
  Yuan Wang$^{1}$\footnotemark[1]\\
  Yan Zhang$^2$\footnotemark[3],\quad 
  Ziyi Chen$^2$,\quad 
  Zhihang Tang$^3$\footnotemark[2],\quad 
  Wei Xu$^2$\footnotemark[3],\quad 
  Zuozhu Liu$^{1}$\footnotemark[2]\\[10pt]
  $^1$Zhejiang University\quad $^2$ByteDance \quad $^3$Zhejiang Lab\\
  \texttt{\{jianhui1.24,zuozhuliu\}@intl.zju.edu.cn}
}

\begin{document}

        

\maketitle

\begingroup \renewcommand{\thefootnote}{\fnsymbol{footnote}} \setcounter{footnote}{0} \footnotetext[1]{Equal contributions.} \footnotetext[2]{Corresponding author.} \footnotetext[3]{Project Leader.} \footnotetext[4]{Work done during internship at ByteDance.} \endgroup

\begin{figure*}[t]
\centering
  \includegraphics[width=1\textwidth]{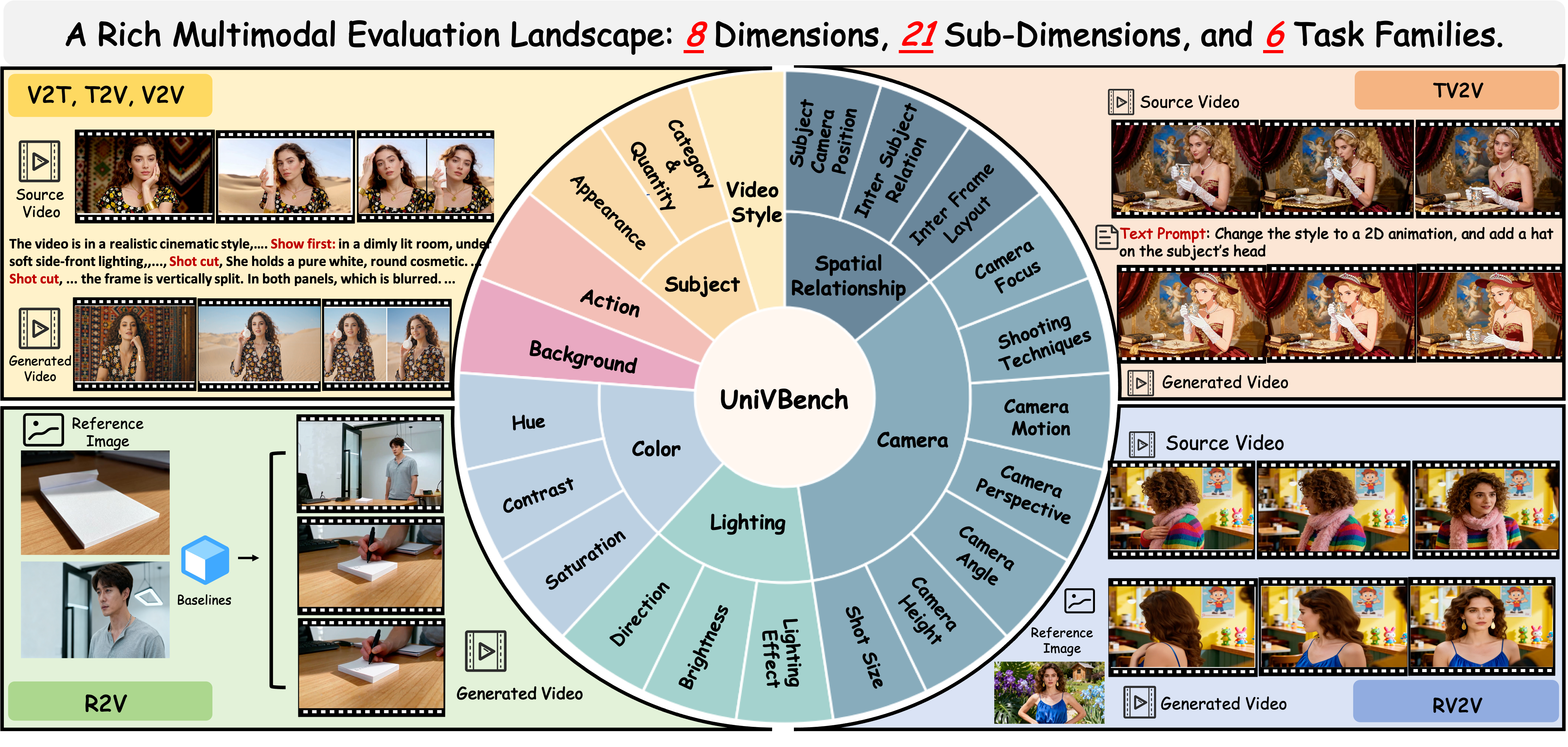}
  \caption{Overview of the \texttt{UniVBench} evaluation setting across 8 Dimensions, 21 Sub-Dimensions, and 6 Tasks. Given a source video, T2V synthesizes a video using its ground-truth caption, while V2V reconstructs the video based solely on the model’s self-generated understanding text, enabling a direct diagnosis of perception–generation coupling. UniVBench supports six unified tasks—video captioning (\textbf{V2T}), text-to-video generation (\textbf{T2V}), reference-image video generation (\textbf{R2V}), text-instruction video editing (\textbf{TV2V}), reference-image video editing (\textbf{RV2V}), and video reconstruction (\textbf{V2V}).
}
  \label{fig:teaser}
\end{figure*}

\begin{abstract}

 Video foundation models aim to integrate video understanding, generation, editing, and instruction following within a single framework, making them a central direction for next-generation multimodal systems. However, existing evaluation benchmarks remain fragmented and limited in scope, as they each target a single task, rely on task-specific metrics, and typically use short or simple video clips. As a result, they do not capture the unified  capabilities that these models are designed to deliver. To address this gap, we introduce UniVBench, a  benchmark purpose-built for evaluating  video foundation models across four core abilities: video understanding, video generation, video editing, and a newly proposed task, video reconstruction, which assesses how faithfully a model can reproduce video content it has  encountered. Our benchmark substantially expands the complexity of evaluation by incorporating 200 high-quality, diverse and multi-shot videos, each paired with detailed captions, multi-format editing instructions, and reference images. All videos are human-created and carefully validated, offering richer cinematic information than prior benchmarks. In addition, we develop a unified agentic evaluation system (UniV-Eval) that standardizes prompting, instruction parsing, and scoring across all tasks, enabling fair, scalable, and reproducible comparisons of unified video models. By grounding evaluation in instruction-based multi-shot video tasks, UniVBench provides the first framework for measuring the integrated capabilities that video foundation models aim to achieve. Extensive human annotations ensure our evaluation aligns with human judgment, enabling rigorous assessment and accelerating progress toward robust video intelligence. Code and datasets are available at \url{https://github.com/JianhuiWei7/UniVBench}
\end{abstract}

\section{Introduction}
\label{sec:intro}

Video foundation models have recently emerged as a promising direction for next-generation multimodal systems. These models aim to integrate understanding and generation within a single architecture. However, current approaches remain fundamentally separated. Generation-focused systems~\cite{Bao2024ViduAH, Polyak2024MovieGA, Ma2025StepVideoT2VTR,wan2025wan,gao2025seedance,3-zhang2025show, Wiedemer2025VideoMA} excel at synthesis but cannot reason about video content, while understanding models~\cite{5-zhang2025videollama,6-li2024llava,7-fang2024vila, Zhu2025InternVL3EA, Bai2025Qwen25VLTR, Guo2025Seed15VLTR,Achiam2023GPT4TR, jiang2025hulu, jiang2024med} achieve strong perception but cannot generate videos. Emerging unified architectures~\cite{8-team2024chameleon,9-zhoutransfusion,10-xieshow,11-wang2024emu3,12-ye2025unic,wu24next,13-chen2025janus,14-deng2025emerging,15-liao2025mogao,tang2024codi,tan2025omni,16-wei2025univideo} attempt to bridge this divide by integrating LLMs with visual tokenizers and video decoders, enabling both video understanding and generation in response to instructions.

Despite these architectural advances, a critical question remains: \textit{does unification actually improve performance across the full spectrum of video tasks?} Current benchmarks cannot answer this due to two fundamental limitations. First, existing datasets are task-specific and cannot support unified evaluation. 
As shown in Table~\ref{tab:longbench_comparison},  most video understanding benchmarks~\cite{zhou2018towards, wang2019vatex, chai2024auroracap, liu2025shotbench} use copyrighted web videos that may contaminate evaluation data and lack the instructions needed for generation and editing tasks. Video generation benchmarks~\cite{chivileva2023measuring,huang2024vbench,li2024genai, zhang2025benchmarking, kou2024subjective, zheng2025vbench,zhang2025q, wang2025love}  focus  on text-to-video synthesis without supporting understanding or editing evaluation. Video editing benchmarks~\cite{wu2023cvpr,feng2024ccedit, singer2024video, ye2025unic, jiang2025vace, li2025five}  remain limited to single-shot scenarios, lacking the multi-shot content. Beyond task coverage, existing benchmarks also exhibit fragmented evaluation of cinematic qualities. As shown in Table~\ref{tab:dimensions_comparison}, understanding benchmarks like AuroraCap~\cite{chai2024auroracap} emphasize subject detection and camera motion but ignore style and spatial relationships; generation benchmarks like VBench~\cite{huang2024vbench} evaluate subjects and actions but lack systematic lighting and color assessment; editing benchmarks like TGVE~\cite{wu2023cvpr} focus on subject preservation but omit background and spatial coherence. No existing benchmark systematically evaluates cinematic dimensions across all video tasks.


Second, evaluation metrics are fundamentally fragmented. As shown in Table~\ref{tab:metrics_comparison}, understanding uses reference-based measures, generation uses distributional metrics that often disagree, and editing requires ad hoc metric combinations~\cite{wuveditbench,zhang2018unreasonable,dosovitskiy2015flownet}. This fragmentation limits cross-task comparison. Moreover, a single scalar score severely limits interpretability, obscuring nuanced trade-offs between a model’s strengths and weaknesses, and consequently failing to provide actionable feedback to the training phase. High-quality evaluation is inherently complex, multifaceted, and dynamic. Fixed evaluation criteria may not generalize across diverse video properties: some videos emphasize faithful reconstruction of visual entities, while others prioritize narrative coherence over instance-level fidelity.

We introduce UniVBench, the first unified benchmark designed to evaluate video foundation models across their full capability spectrum. The benchmark comprises 200 high-quality, multi-shot videos, each with rich annotations including detailed captions, multi-format editing instructions, and reference images. Crucially, all content is human-created and copyright-free, enabling fair evaluation of editing, reconstruction, and instruction-following without legal or data contamination concerns. We pair the benchmark with a unified agentic evaluation system (UniV-Eval) that standardizes prompting, instruction parsing, and multi-dimensional scoring across all tasks. This provides consistent, interpretable metrics that enable direct cross-model and cross-task comparison, while supporting fine-grained attribution of errors to perception versus generation components.

Our work makes three key contributions: (1) The first multi-shot video dataset specifically designed for unified evaluation, free of copyright and contamination issues; (2) a unified agentic evaluation system that enables  measurement across understanding, generation, editing, and reconstruction; and (3) a principled framework for attributing model capabilities and failures across the perception-generation spectrum. By aligning evaluation with the goals of unified video modeling, our benchmark establishes a foundation for measuring progress toward general-purpose, instruction-following video intelligence.




\begin{table}[t]
\centering
\small
\setlength{\tabcolsep}{5pt}
\renewcommand{\arraystretch}{1.0}
\resizebox{0.47\textwidth}{!}{
\begin{tabular}{lccc}
\toprule
\multicolumn{4}{l}{\textbf{Tasks} : \ding{172}: V2T \ \ \ \ \  \ding{173}: T2V \ \ \ \ \ \ding{174}: R2V \ \ \ \ \ \ding{175}: TV2V \ \ \ \ \ \ding{176}: RV2V \ \ \ \ \ \ding{177}: V2V} \\
\hline
\hline
\textbf{Benchmark} & 
\textbf{Multi-task} & 
\textbf{Multi-shot} & 
\textbf{Copyright Issue} \\
\midrule
\multicolumn{4}{l}{\textit{Benchmarks for Video Understanding}} \\
\midrule
M-VAD~\cite{torabi2015using} & \ding{172} & \xmark   & Yes  \\
MPII-MD~\cite{venugopalan2015translatingvideosnaturallanguage} & \ding{172} & \xmark  & Yes  \\
MSR-VTT~\cite{xu2016msr} & \ding{172} & \cmark  &Potential   \\
Charades~\cite{sigurdsson2016hollywood} & \ding{172} & \xmark  &Yes \\
ActivityNet~\cite{krishna2017dense} & \ding{172} & \cmark  &Yes   \\
Youcook2~\cite{zhou2018towards} &  \ding{172} & \xmark  & Yes\\
VATEX~\cite{wang2019vatex} & \ding{172} & \xmark  & Yes \\
AuroraCap~\cite{chai2024auroracap} & \ding{172} & \xmark  & Potential \\
ShotBench~\cite{liu2025shotbench} & \ding{172} & \xmark  & Yes \\
\midrule
\multicolumn{4}{l}{\textit{Benchmarks for Video Generation}} \\
\midrule
EvalCrafter~\cite{liu2024evalcrafter} & \ding{173} & \xmark  & NA  \\
FETV~\cite{liu2023fetv} & \ding{173} & \xmark & NA  \\
MQT~\cite{chivileva2023measuring} & \ding{173} & \xmark & NA  \\
VBench~\cite{huang2024vbench} & \ding{173} & \xmark  & NA  \\
GenAIBench~\cite{li2024genai} & \ding{173} & \xmark  & NA  \\
LGVQ~\cite{zhang2025benchmarking} & \ding{173} & \xmark  & NA  \\
T2VQA-DB~\cite{kou2024subjective} & \ding{173} & \xmark  & NA  \\
AIGVQA-DB~\cite{wang2025aigv} & \ding{173} & \xmark  & NA  \\
VBench2.0~\cite{zheng2025vbench} & \ding{173} & \xmark  & NA  \\
Q-Eval~\cite{zhang2025q} & \ding{173} & \xmark  & NA  \\
AIGVE-60K~\cite{wang2025love} & \ding{173} & \xmark  & NA  \\
\midrule
\multicolumn{4}{l}{\textit{Benchmarks for Video Editing}} \\
\midrule
CCEdit~\cite{feng2024ccedit} & \ding{175} &\xmark & No  \\
TGVE~\cite{wu2023cvpr} & \ding{175}  & \xmark & No  \\
TGVE+~\cite{singer2024video} & \ding{175}  & \xmark & No  \\
VE-Bench~\cite{sun2025ve} & \ding{175} &  \xmark & No \\
UNIC~\cite{ye2025unic} & \ding{176}  &  \xmark & Potential \\
VACE-Bench~\cite{jiang2025vace} & \ding{174} \ding{175} \ding{176} &  \xmark & Potential \\
FIVE~\cite{li2025five} &  \ding{175} &  \xmark & Potential \\
\midrule
\rowcolor{gray!15}
\textbf{UniVBench} & \ding{172}$\sim$\ding{177} & \textbf{\cmark} & No \\
\bottomrule
\end{tabular}
}
\caption{Comparison of benchmarks for video understanding, generation and editing. \textbf{Multi-task} shows the applicable tasks of the benchmark. \textbf{Multi-shot} indicates that whether the video source and text annotations have multi-shot content. \textbf{Copyright Issue} indicates whether the video sources in the dataset are editable without copyright issue. }
\label{tab:longbench_comparison}
\end{table}

\section{Related Work}

\subsection{Video Foundation Models}
Early progress in video foundation models emerged from text-to-video generation, where diffusion-based frameworks such as ModelScopeT2V~\cite{1-wang2023modelscope}, LAMP~\cite{wu2024lamp}, CogVideoX~\cite{2-yang2024cogvideox}, Vidu~\cite{Bao2024ViduAH} and Wan~\cite{wan2025wan} achieved high-fidelity video synthesis, while autoregressive models like Show-1~\cite{3-zhang2025show} and Emu2~\cite{4-sunemu} introduced unified visual tokens for more controllable generation. However, these approaches are inherently one-directional, focusing on synthesis without true multimodal reasoning. In parallel, video-to-text understanding models including VideoLLaMA3~\cite{5-zhang2025videollama}, LLaVA-OneVision~\cite{6-li2024llava}, and VILA$^{2}$~\cite{7-fang2024vila} extended large multimodal encoders to interpret temporal content and perform grounded question answering. Despite strong perception ability, these encoder-based methods remain limited to understanding and cannot generate or reconstruct visual dynamics, leaving a clear separation between perception- and generation-oriented paradigms. To close this divide, unified video foundation models have recently emerged, seeking to integrate understanding, generation, and editing within a single architecture. Representative works such as Chameleon~\cite{8-team2024chameleon}, Transfusion~\cite{9-zhoutransfusion}, Show-o~\cite{10-xieshow}, Emu3~\cite{11-wang2024emu3}, and UNIC~\cite{12-ye2025unic} jointly train autoregressive and diffusion objectives to unify decoding across text and visual tokens. Further developments like NExT-GPT~\cite{wu24next}, Janus-Pro~\cite{13-chen2025janus}, BAGEL~\cite{14-deng2025emerging}, Mogao~\cite{15-liao2025mogao}, CoDi-2~\cite{tang2024codi}, Omni-Video~\cite{tan2025omni}, and UniVideo~\cite{16-wei2025univideo} incorporate large language models with 3D visual tokenizers and causal VAEs to achieve bidirectional reasoning over text, image, and video modalities.

\begin{table}[htbp]
\centering
\setlength{\tabcolsep}{5pt}
\renewcommand{\arraystretch}{1.2}
\resizebox{0.47\textwidth}{!}{
\begin{tabular}{lcccccccc}
\toprule
\textbf{Benchmarks} & \textbf{Style} & \textbf{Subject} & \textbf{Action} & \textbf{Backg.} & \textbf{Lighting} & \textbf{Color} & \textbf{\makecell{Spatial\\Relationship}} & \textbf{Camera} \\
\midrule
\textbf{AuroraCap} & \xmark & \textbf{\cmark} & \xmark & \textbf{\cmark} & \xmark & \xmark & \xmark & \textbf{\cmark} \\
\textbf{VGenEval} & \textbf{\cmark} & \textbf{\cmark} & \textbf{\cmark} & \textbf{\cmark} & \textbf{\cmark} & \xmark & \textbf{\cmark} & \textbf{\cmark} \\
\textbf{ShotBench} & \xmark & \xmark & \xmark & \xmark & \textbf{\cmark} & \xmark & \xmark & \textbf{\cmark} \\
\textbf{FETV} & \xmark & \textbf{\cmark} & \textbf{\cmark} & \textbf{\cmark} & \textbf{\cmark} & \xmark & \xmark & \xmark \\
\textbf{VBench} & \xmark & \textbf{\cmark} & \textbf{\cmark} & \textbf{\cmark} & \xmark & \textbf{\cmark} & \textbf{\cmark} & \xmark \\
\textbf{VBench2.0} & \xmark & \textbf{\cmark} & \textbf{\cmark} & \textbf{\cmark} & \xmark & \xmark & \textbf{\cmark} & \textbf{\cmark} \\

\textbf{Charades} & \xmark & \textbf{\cmark} & \textbf{\cmark} & \xmark & \xmark & \xmark & \xmark & \xmark \\

\textbf{YouCook} & \xmark & \xmark & \textbf{\cmark} & \xmark & \xmark & \xmark & \xmark & \xmark \\

\textbf{MPII-MD} & \xmark & \textbf{\cmark} & \textbf{\cmark} & \xmark & \xmark & \xmark & \xmark & \xmark \\

\textbf{EvalCrafter} & \textbf{\cmark}  & \textbf{\cmark} & \xmark & \textbf{\cmark} & \xmark & \xmark & \xmark & \textbf{\cmark} \\

\textbf{TGVE} & \textbf{\cmark}  & \textbf{\cmark}  & \xmark & \textbf{\cmark} & \xmark & \xmark & \xmark & \xmark  \\

\textbf{TGVE} & \textbf{\cmark}  & \textbf{\cmark}  & \xmark & \textbf{\cmark} & \xmark & \xmark & \xmark & \xmark  \\

\textbf{UNIC} & \textbf{\cmark}  & \textbf{\cmark}  & \xmark   & \xmark & \xmark & \xmark   & \xmark & \textbf{\cmark}    \\

\textbf{FiVE} & \xmark  & \textbf{\cmark}  & \textbf{\cmark}   & \xmark & \xmark & \textbf{\cmark}   & \xmark & \textbf{\cmark}    \\

\textbf{VE-Bench} & \xmark  & \textbf{\cmark}  & \textbf{\cmark}   & \xmark & \xmark & \xmark   & \xmark & \xmark    \\

\textbf{CC-Edit} & \xmark  & \textbf{\cmark}  & \textbf{\cmark}   & \textbf{\cmark} & \xmark & \xmark   & \xmark & \textbf{\cmark}    \\
\midrule
\rowcolor{gray!15}
\textbf{UniVBench} & \textbf{\cmark} & \textbf{\cmark} & \textbf{\cmark} & \textbf{\cmark} & \textbf{\cmark} & \textbf{\cmark} & \textbf{\cmark} & \textbf{\cmark} \\
\bottomrule
\end{tabular}
}
\caption{Comparison of cinematic dimensions across different video evaluation benchmarks.}
\label{tab:dimensions_comparison}
\end{table}

\subsection{Video Benchmark}


Early video understanding benchmarks like M-VAD~\cite{torabi2015using}  and MPII-MD~\cite{venugopalan2015translatingvideosnaturallanguage} focused on single-shot video captioning with limited scale. Larger benchmarks such as MSR-VTT~\cite{xu2016msr} and ActivityNet~\cite{krishna2017dense} expanded dataset size and introduced multi-shot content, enabling evaluation of temporal reasoning and long-form video understanding. More recent efforts like AuroraCap~\cite{chai2024auroracap} and ShotBench~\cite{liu2025shotbench}  have improved annotation quality and introduced shot-level analysis. However, these benchmarks primarily use web-scraped videos, raising potential copyright and data contamination concerns when evaluating models trained on large-scale internet data.

With the emergence of text-to-video models, specialized generation benchmarks have been developed. Early works like FETV~\cite{liu2023fetv} and MQT~\cite{chivileva2023measuring} introduced basic quality metrics, while VBench~\cite{huang2024vbench} established a comprehensive evaluation framework with 16 dimensions covering quality, semantics, and temporal consistency. Subsequent benchmarks like GenAIBench~\cite{li2024genai}, LGVQ~\cite{zhang2025benchmarking} , T2VQA-DB~\cite{kou2024subjective}   expanded evaluation to include subjective quality assessment and diverse generation scenarios. Recent efforts like VBench2.0~\cite{zheng2025vbench}, Q-Eval~\cite{zhang2025q} , and AIGVE-60K~\cite{wang2025love} have scaled up evaluation with larger test sets and more refined metrics. However, all existing generation benchmarks focus exclusively on text-to-video synthesis, lacking support for reference-guided generation or editing.

Video editing benchmarks have primarily focused on instruction-following capabilities. CCEdit~\cite{feng2024ccedit}  and TGVE~\cite{wu2023cvpr} pioneered text-guided editing assessment, measuring both editing accuracy and video quality preservation. TGVE+~\cite{singer2024video}  and VE-Bench~\cite{sun2025ve} extended evaluation to more complex editing scenarios with fine-grained metrics. Recent works like UNIC~\cite{ye2025unic} introduced reference image-based editing, while VACE-Bench~\cite{jiang2025vace} attempted to unify multiple editing modalities. FIVE~\cite{li2025five}  provided adversarial test cases for robust editing evaluation. Despite these advances, existing editing benchmarks are limited to single-shot videos and do not support multi-shot content, which is essential for evaluating real-world video editing capabilities.

Overall, current benchmarks suffer from three critical limitations: they are task-specific, restricted to single-shot scenarios, and understanding benchmarks often use copyrighted content that may contaminate evaluation. Our UniVBench addresses all these limitations by providing the first multi-task, multi-shot benchmark with copyright-free content, enabling comprehensive evaluation across the full spectrum of video tasks.


\begin{table}[t]
\centering
\setlength{\tabcolsep}{5pt}
\renewcommand{\arraystretch}{1.0}
\resizebox{0.47\textwidth}{!}{
\begin{tabular}{lcccc} 
\toprule
\multicolumn{5}{l}{\textbf{Tasks} : \ding{172}: V2T \ \ \ \ \  \ding{173}: T2V \ \ \ \ \ \ding{174}: R2V \ \ \ \ \ \ding{175}: TV2V \ \ \ \ \ \ding{176}: RV2V \ \ \ \ \ \ding{177}: V2V} \\
\hline
\hline
\textbf{Evaluation Method} & \textbf{Fine-grained} & \textbf{Multi-shot} & \textbf{Multi-dimension} & \textbf{Task} \\ 
\midrule
BLEU~\cite{papineni2002bleu}   & \xmark  & \cmark & \xmark & \ding{172}  \\
CLIPScore~\cite{hessel2021clipscore}   & \xmark  & \xmark & \cmark & \ding{173}  \\
CIDEr~\cite{vedantam2015cider}  & \xmark & \cmark & \xmark  & \ding{172}  \\
FVD~\cite{unterthiner2019fvd}  & \xmark & \xmark  & \cmark  & \ding{173} \\
CLIPSIM~\cite{kou2024subjective} &  \xmark & \xmark & \cmark  & \ding{173} \\
LPIPS~\cite{zhang2018unreasonable}   & \xmark & \xmark  & \xmark  & \ding{173}  \\
LLM-as-a-Judge~\cite{li2025generation} & \cmark   & \cmark  & \xmark  & \ding{172}$\sim$\ding{177}  \\
\midrule
\rowcolor{gray!15}
\textbf{UniV-Eval}  & \cmark   & \cmark & \cmark & \ding{172}$\sim$\ding{177} \\
\bottomrule
\end{tabular}
}
\caption{Comparison of core capabilities across existing evaluation metrics and our proposed agent-based evaluation system. ``-'' indicates the metric is not applicable to this dimension.} 
\label{tab:metrics_comparison}
\end{table}

\subsection{Video Evaluation Methods}

Video evaluation has traditionally relied on task-specific metrics that lack the flexibility required for unified assessment. For video understanding, BLEU~\cite{papineni2002bleu} and CIDEr~\cite{vedantam2015cider} measure n-gram overlap between generated and reference captions, providing coarse-grained quality scores but failing to capture semantic nuances or fine-grained errors. For video generation, metrics like FVD~\cite{unterthiner2019fvd} assess distributional similarity between generated and real videos, while CLIPScore~\cite{hessel2021clipscore}  and CLIPSIM~\cite{kou2024subjective} measure semantic alignment between videos and text prompts. LPIPS~\cite{zhang2018unreasonable} evaluates perceptual similarity for frame-level reconstruction. For video editing, evaluation typically combines multiple metrics, using LPIPS~\cite{zhang2018unreasonable} for background preservation, CLIPScore~\cite{hessel2021clipscore}  for instruction alignment, and frame-by-frame comparisons for temporal consistency. However, these metrics are fundamentally limited: they operate at the video or dataset level without fine-grained error attribution, most cannot handle multi-shot videos, and each is designed for a specific task, limiting cross-task comparison.

Recent work~\cite{huang2024vbench, zheng2025vbench, wang2025love, li2025generation} has explored LLM-as-a-Judge approaches that use vision-language models to provide qualitative assessments across multiple tasks. While these methods offer flexibility and can handle diverse inputs including editing scenarios, they typically produce single overall scores without multi-dimensional analysis, limiting their diagnostic value for model development. Overall, no existing evaluation method simultaneously provides fine-grained analysis, multi-shot support, multi-dimensional scoring, and cross-task applicability. Our agentic evaluation system addresses these limitations by decomposing evaluation into interpretable dimensions, providing shot-level attribution, and maintaining consistent criteria across all video tasks.
\section{UniVBench}


\subsection{Dataset Construction}

\paragraph{Video Synthesis.} To ensure comprehensive cinematic coverage, we adopt eight fundamental dimensions from prior works ~\cite{yang2025videogenevalagentbasedvideogeneration, li2024genai, gao2025seedance, huang2024vbench} and extend them with 21 fine-grained sub-dimensions (Figure~\ref{fig:teaser}): style, subject (category, quality, appearance), action, background, camera (focus, shot size, motion, perspective, angle, height, techniques), lighting (direction, brightness, effect), color (hue, contrast, saturation), and spatial relationships(iter-frame/subject layout, camera-subject position). We pre-classify categories for each sub-dimension (e.g., styles: realistic, animation, 2D; camera movements: static, zoom, pan, tracking; lighting: daylight, golden hour, studio).

We recruit 15 professional experts with video production backgrounds who receive detailed training on our dimension taxonomy and annotation guidelines. For script writing, annotators sample random category combinations and compose detailed narratives specifying all dimension attributes shot-by-shot. Each multi-shot script must maintain narrative coherence across shots while covering diverse dimension values. Scripts undergo peer review where a second annotator verifies dimension coverage and coherence before generation.

We generate videos using top commercial APIs (Hailuo, Kling, Veo3) and apply three-stage human-in-the-loop filtering: (1) automated pre-filtering removes watermarks and IP content via vision-language models, (2) three trained reviewers independently verify each video's adherence to script specifications across all eight dimensions, accepting only videos with unanimous agreement, (3) quality specialists inspect for artifacts, unnatural motion, and temporal inconsistencies. Videos failing any stage are regenerated or discarded. On average, each video undergoes 2.3 generation attempts before approval. This rigorous process yields 100 single-shot and 100 multi-shot videos (avg. 3.72 shots).

\begin{figure*}[t]
\centering
\includegraphics[width=0.98\textwidth]{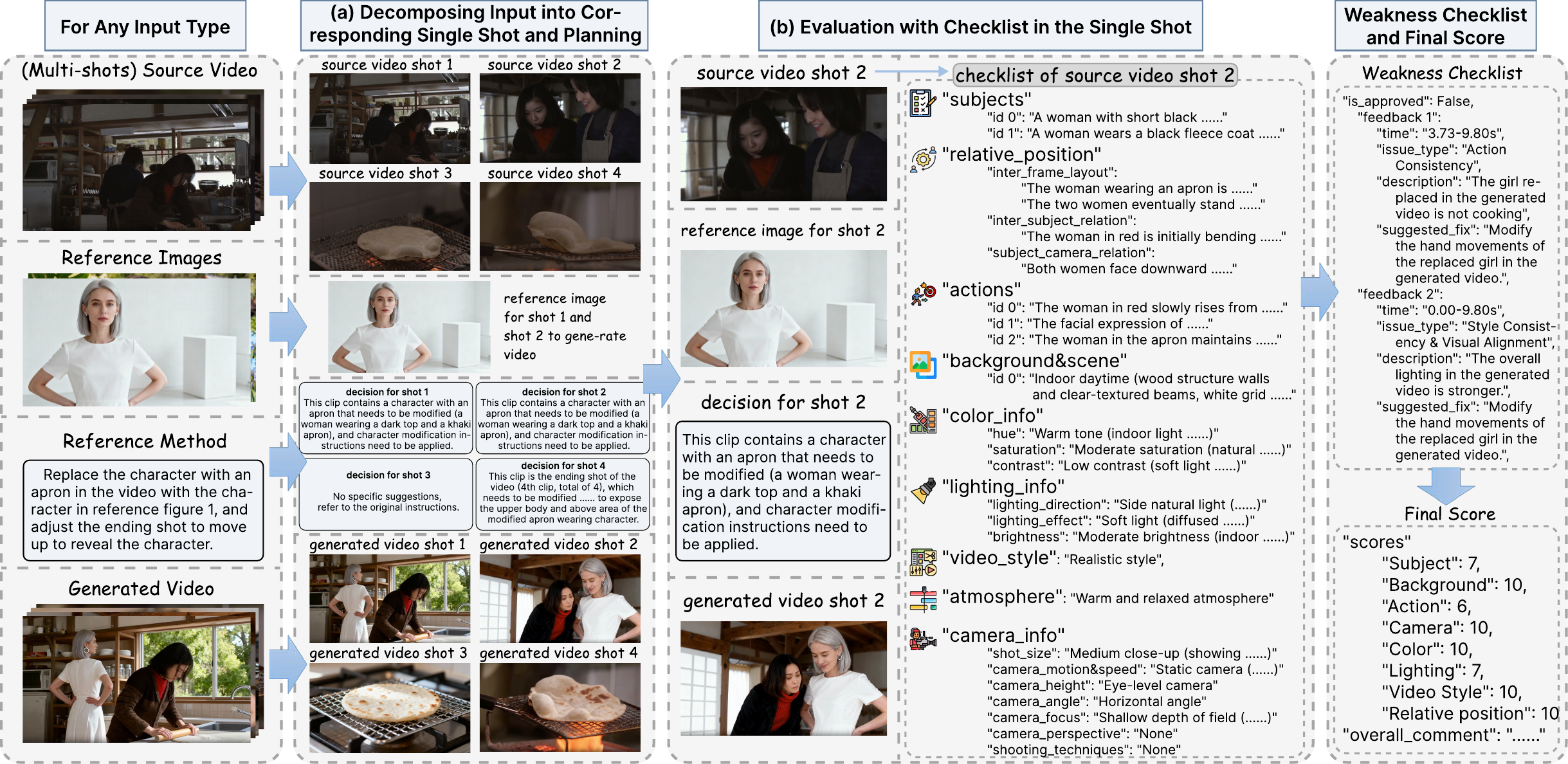}
  \caption{Workflow of UniV-Eval. The system accepts arbitrary inputs within a task setting and performs dynamic evaluation after planning and decomposition. The final results are delivered as a fine-grained checklist, providing traceable feedback for training optimization.
}
  \label{fig:evaluation case}
\end{figure*}

\paragraph{Detailed Captioning.} We generate dimension-complete ground-truth captions using Gemini 2.5 Pro through: (1) dimension-wise extraction for all eight dimensions and sub-dimensions, (2) synthesis into coherent, shot-level descriptions. Three annotators then independently verify each caption against the source video, checking dimension completeness and temporal accuracy. GPT-4o provides additional automated verification by cross-checking factual claims. Captions with any disagreement undergo collaborative review where annotators discuss discrepancies and produce corrected versions. Each caption is revised an average of 1.8 times before finalization.

\noindent \textbf{Reference Images.}
To construct a diverse reference image sets for R2V, RV2V tasks, we generate high-fideility reference images using Gemini 2.5 Flash Image (Nano Banana) and Seedream4.0 ~\cite{seedream2025seedream40nextgenerationmultimodal}. We firstly define three type of reference images: subject, style, and scene. For subject, we also define human subjects, animals, non-living objects(clothes, paper, etc.,). For style, it mainly covers 6 major styles: animation(2D, 3D), real(cinematic style, ), arts(Japanese ukiyoe style), sci-fi(cyberpunk style, wasteland style), dressing (rococo, lolita), and materials(clay animation style, building block style). For background, it is divied into natural (with different seasons, weather and time), human crafted (street, buildings), and vritual (magic library) scenes. The generated images are also carefully picked to ensure quality and prevent any infringement. Finally, 864 unique and diverse images are created for reference image related tasks.

\noindent \textbf{Video2Video Reconstruction.}
We innovatively propose a new task, Video2Video reconstruction, to evaluate the performance of unified models in both understanding and generation tasks. Specifically, this task first requires the model to understand a video and generate corresponding detailed captions, then reconstruct the video based on the generated text. By directly comparing the reconstructed video with the original one, we can assess the unified model’s capabilities in understanding and generation. A high-quality unified model should first generate excellent captions through understanding, and second produce high-quality videos based on text. Failure in either task will result in a significant discrepancy between the reconstructed video and the origin.

\subsection{UniV-Eval}
We first introduce the evaluation tasks and corresponding evaluation strategies encompassed by our proposed unified evaluation system.
Existing evaluation approaches typically assess the performance of video understanding and generation models on isolated tasks, in a decoupled manner, lacking a unified and integrated evaluation framework. Meanwhile, these methods often oversimplify the evaluation process, which leads to several potential risks:
First, producing a single scalar score can severely limit interpretability, failing to reveal fine-grained distinctions between the model’s strengths and weaknesses. As a result, evaluations based solely on aggregate metrics make it challenging to provide actionable feedback for refinement during training.
Second, assessing high-quality generation inherently involves complex, multifaceted, and dynamic dimensions. Fixed evaluation criteria may fail to accommodate diverse video attributes. For instance, some test cases emphasize the faithful reconstruction of visual instances, while others prioritize narrative coherence over instance-level fidelity.

Therefore, in contrast to performing single-valued and fixed-dimension evaluations of video generation quality, we propose a dynamically adaptive, fine-grained agentic system UniV-Eval that decomposes overall ``generation performance" into a set of interpretable, multidimensional checklists. This design enables a more comprehensive and diagnostic assessment of model capability beyond conventional single-score evaluations. Specifically, as a complementary component to UniVBench, our evaluation system centers on the user instruction and standardizes the prompting and instruction parsing procedures across tasks. Given any input (source video, reference image, and reference text), the system enables the evaluation of any output (including both video and text) in a unified and consistent manner.

\paragraph{Decomposing and Planning.} Due to the current limitation on model generation length, a long video $\mathcal{V}$ is mechanically segmented into multiple clips $\mathcal{V} = \{\bm{c}_i\}_{i=1}^C$ for both generation and evaluation.
As illustrated in Figure~\ref{fig:evaluation case} (a), the proposed agent system UniV-Eval first decomposes each clip-level input into shot-level units for evaluation. Specifically, the multi-shots video is segmented as $V=\{v_1,v_2,...,v_n\}$ using PySceneDetect\footnote{A tool designed to extract the minimal sub-shots from multi-shot videos.}, thereby determines the number $n$ of shot-level units in a single clip.

Meanwhile, the \texttt{shot\_classification} agent aligns the reference images $I$ and initial user instruction $T$ with their corresponding shots: $I =\{i_1,i_2,...,i_n\}$ and $T =\{t_1,t_2,...,t_n\}$
, resulting in a set of shot-level inputs $(v, i, t)$ that serve as the foundation for subsequent evaluation. Notably, in the proposed system, all input modalities are optional, allowing flexible combinations of inputs depending on the evaluation scenario.

\paragraph{Shot-level Fine-grained Evaluation.}
Let the output of the tested model as o\_1, combining with the input tuple $(v, i, t)$, we invoke the \texttt{shot\_evaluation} agent to perform assessment. To ensure fine-grained scene and visual understanding at the shot level, we design nine major category groups: subject, relative\_position, actions, background\&scene, color\_info, lighting\_info, video\_style, atmosphere and camera\_info~\cite{studiobinder1,studiobinder2,studiobinder3,studiobinder4,studiobinder5}, as shown in the checklist of Figure~\ref{fig:evaluation case} (b). Each major category is further decomposed into specific, interpretable subcategories (21 in total) that the \texttt{shot\_evaluation} agent scores and reports, enabling diagnostic, fine-grained feedback at the shot level.

The \texttt{shot\_evaluation} agent performs per-category comparisons between the model output o\_1 and the input tuple $(v, i, t)$, producing a structured weakness checklist that highlights fine-grained deficiencies useful for targeted training optimization. This checklist is then forwarded to an \texttt{evaluation\_score agent}, which aggregates the diagnostic signals and issues final scores along six evaluation dimensions for quantitative performance comparison.

\section{Experiments}

\begin{table*}[htbp]
\centering
\resizebox{\linewidth}{!}{%
\begin{tabular}{l|c|cccccccc|c}
\toprule
\textbf{Task} & \textbf{Models} & \textbf{Subject} & \textbf{Background} & \textbf{Action} & \textbf{Camera} & \textbf{Color} & \textbf{Lighting} & \textbf{Video Style} & \textbf{Relative Position} & \textbf{Average} \\
\midrule
\centering
\multirow{7}{*}{Understanding (V2T)} 
& Gemini 2.5 Pro$^{\ddag}$ & 54.4\% & 57.8\% & 27.0\% & 54.1\% & 65.8\% & 63.8\% & 65.4\% & 44.4\% & 54.1\% \\
& Seed 1.6$^{\ddag}$ & 46.3\% & 46.9\% & 22.3\% & 42.8\% & 54.6\% & 49.7\% & 45.9\% & 33.8\% & 42.8\% \\
& Qwen3-VL-30B$^{\S}$ & 15.8\% & 14.6\% & 6.4\% & 17.6\% & 25.3\% & 27.1\% & 24.1\% & 9.8\% & 17.6\% \\
& AuroraCap$^{\S}$ & 16.7\% & 14.4\% & 6.9\% & 17.8\% & 23.9\% & 26.4\% & 25.2\% & 10.8\% & 17.8\%\\
& Tarsier2$^{\S}$ & 34.3\% & 25.0\% & 32.7\% & 20.9\% & 7.7\% & 4.4\% & 20.1\% & 21.9\% & 21.9\% \\
& Showo-2$^{\S}$ & 25.9\% & 22.2\% & 10.6\% & 16.3\% & 13.7\% & 11.3\% & 18.3\% & 12.3\% & 16.3\% \\
\midrule

\multirow{5}{*}{Generation (T2V)} 
& Seedance-1.0-Pro$^{\ddag}$ & 68.8\% & 65.2\% & 74.3\% & 76.5\% & 84.8\% & 83.4\% & 76.8\% &  91.6\% & 77.9\% \\
& Wan2.2-14B$^{\S}$ &  70.0\% & 79.7\% & 62.1\% & 72.2\% & 81.6\% & 69.2\% & 79.9\% & 90.0\% & 74.9\% \\
& CoDi-2$^{\S}$ & 6.1\% & 15.6\% & 25.0\% & 44.7\% & 54.6\% & 55.7\% & 37.1\% &  83.4\% & 40.1\% \\
& Omni-Video$^{\S}$ & 45.0\% & 52.6\% & 41.2\% & 49.6\% & 67.6\% & 66.2\% & 60.6\% & 66.5\% & 56.2\% \\
& CogVideoX$^{\S}$ & 42.6\% & 62.9\% & 33.9\% & 61.7\% & 82.0\% & 83.2\% & 77.8\% & 81.3\% & 65.7\%\\
\midrule

\multirow{1}{*}{Generation (R2V)} 
& Seedance-1.0-Lite$^{\ddag}$ & 64.7\% & 68.2\% & 39.8\% & 63.1\% & 75.4\% & 74.2\% & 73.8\% & 74.4\% & 66.7\% \\
\midrule

\multirow{1}{*}{Editing (TV2V)} 
& Wan2.1-VACE-14B$^{\S}$ & 66.3\% & 51.2\% & 45.3\% & 62.5\% & 75.3\% & 68.9\% & 72.5\% & 78.4\% & 65.1\% \\
\midrule

\multirow{1}{*}{Editing (RV2V)} 
& Wan2.1-VACE-14B$^{\S}$ & 53.1\% & 57.3\% & 71.1\% & 70.5\% & 70.1\% & 74.1\% & 64.0\% & 71.2\% & 66.4\% \\
\midrule

\multirow{4}{*}{Reconstruction (V2V)} 
& Omni-Video$^{\S}$ & 20.7\% & 29.1\% & 19.8\% & 71.5\% & 59.8\% & 63.3\% & 37.3\% & 81.6\% & 47.9\% \\
& Wan2.1-VACE-1.5B$^{\S}$ & 7.1\% & 6.9\% & 29.0\% & 69.2\% & 32.7\% & 37.0\% & 17.6\% & 79.5\% & 34.9\% \\
& Wan2.1-VACE-14B$^{\S}$ & 56.4\% & 60.4\% & 68.2\% & 77.5\% & 40.9\% & 66.7\% & 51.3\% & 79.9\% & 62.7\% \\
& CogVideoX-1.5-5B$^{\S}$ & 4.6\% & 6.1\% & 12.7\% & 47.2\% & 15.4\% & 34.8\% & 6.7\% & 37.8\% & 20.7\% \\
\bottomrule
\end{tabular}
}

\footnotesize \textbf{Note:} Model types are separated into: $^{\ddag}$Commercial Models, $^{\S}$Open-Source Models.

\caption{Performance comparison of different baselines on UniVBench, summarizing results over six tasks, across eight dimensions.}
\label{main_table}
\end{table*}


\begin{figure*}[htbp]
\centering
  \includegraphics[width=1\textwidth]{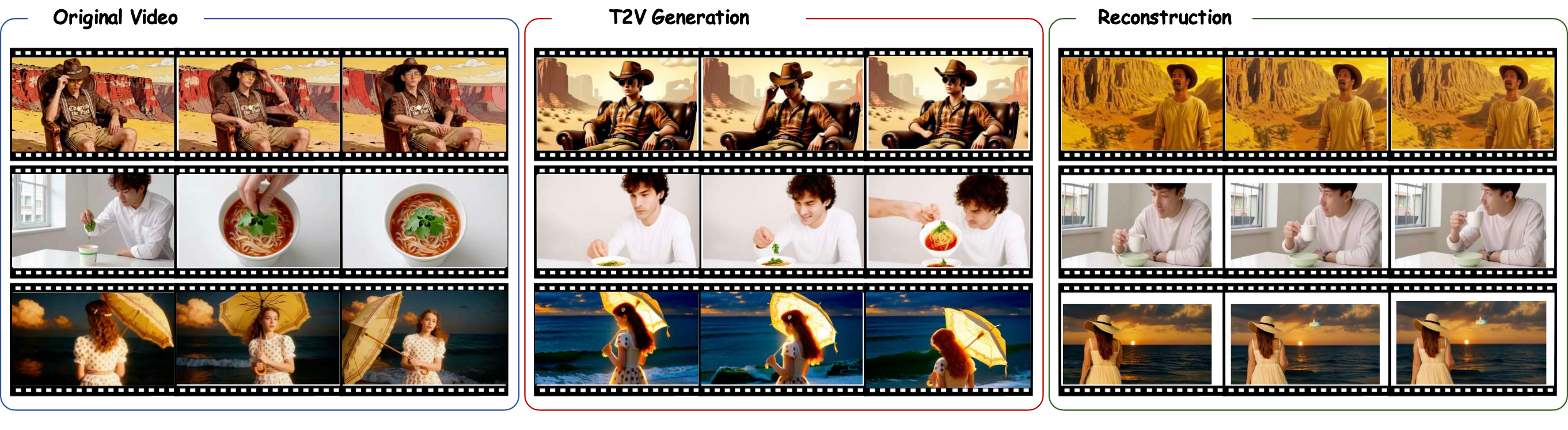}
  \caption{Case Study Analysis of UniVBench in T2V and Reconstruction Task. T2V generation uses the ground truth text of the video, while V2V reconstruction relies on model's understanidng text. The generated videos are selected from OmniVideo
}
  \label{fig:case_study}
\end{figure*}

\subsection{Implementation Details}
For a fair and reproducible comparison, we evaluate all baselines under a unified experimental protocol.
For commercial large multimodal models such as GPT-5, Gemini 2.5 Pro~\cite{comanici2025gemini}, Seed 1.6~\cite{Guo2025Seed15VLTR}, and Seedance-1.0-Lite~\cite{gao2025seedance}, we directly access their official inference APIs released in late 2025. For open-source baselines, including CogVideoX~\cite{2-yang2024cogvideox}, CoDi-2~\cite{tang2024codi}, Omni-Video~\cite{tan2025omni}, Wan2.1-VACE~\cite{wan2025wan}, and other video generation or editing models, we use their official codebases and pre-trained checkpoints. All models use consistent inference settings: 50 DDIM sampling steps, classifier-free guidance scale of 7.5, and native resolution (typically 720×480 for 16:9 videos). All models are executed under consistent settings, including fixed sampling steps, classifier-free guidance scales, and resolution configurations aligned with their default or recommended parameters. When models lack native support for certain tasks, we implement minimal adaptations: for TV2V editing, we concatenate instruction text with source video embeddings; for RV2V editing, we inject reference image features into the diffusion process at intermediate layers. We use Seed-1.6~\cite{Guo2025Seed15VLTR} as the evaluation LLM.




All models receive identical inputs per task: ground-truth captions for T2V generation, source videos and editing instructions for TV2V, reference images and prompts for R2V. For V2V reconstruction, we first generate captions using each model's understanding component (or GPT-4o for generation-only models), then reconstruct using those captions. Video inputs are center-cropped and resized to each model's expected resolution while maintaining aspect ratio. All outputs undergo evaluation by our agentic system using identical prompts, rubrics, and dimension weightings, ensuring differences in scores reflect model capabilities rather than evaluation variance. All experiments are conducted on 8 NVIDIA H100 GPUs 80GB.

\subsection{Main Results}

Our comprehensive evaluation on UniVBench, summarized in Table ~\ref{main_table}, reveals a distinct specialization among current video models, highlighting the performance gap between systems designed for single task versus those for unified tasks.

\paragraph{Task-Specific Leaders.} In the video understanding (V2T) task, Gemini 2.5 Pro demonstrates superior performance with an average score of 54.1\%, significantly outpacing other models. Conversely, unified video models like Showo-2 score (16.3\%) in this domain, showcasing their lack of perceptual reasoning. For text-to-video (T2V) generation, Seedance-1.0-Pro achieves the top score of 77.9\%. For reconstruction (V2V), Wan2.1-VACE-14B delivers the strongest performance, with scores of 62.7\%.

\paragraph{Cross-Dimensional Insights.} A key observation across all tasks is the difficulty models have with the `Action' dimension, which frequently receives the lowest scores, particularly in video understanding. This suggests that accurately interpreting and synthesizing complex temporal dynamics remains a major challenge. In contrast, generative models exhibit greater control over stylistic attributes like `Color', `Lighting', and `Video Style', where they often achieve their highest scores. 

\paragraph{The Unification Gap.} Overall, the results quantitatively indicates that no single model currently excels across the full spectrum of understanding, generation, and editing. The benchmark effectively maps the strengths and weaknesses of existing architectures, providing a clear and necessary baseline to guide future efforts in developing truly unified video foundation models.

\subsection{Analysis}
\paragraph{Reconstruction Case Study.} We present qualitative results in Figure ~\ref{fig:case_study}, where T2V generation leverages the video’s ground truth text and reconstruction relies on the model’s self-derived understanding text. A comparison of these three sets of videos reveals varying degrees of inconsistency. Notably, the V2V task exhibits more pronounced inconsistencies than its T2V counterpart, indicating information transmission loss during the V2T → T2V pipeline. Collectively, these findings clearly highlight the inherent weaknesses of current unified video models.

\begin{figure}[t]
\centering
  \includegraphics[width=0.47\textwidth]{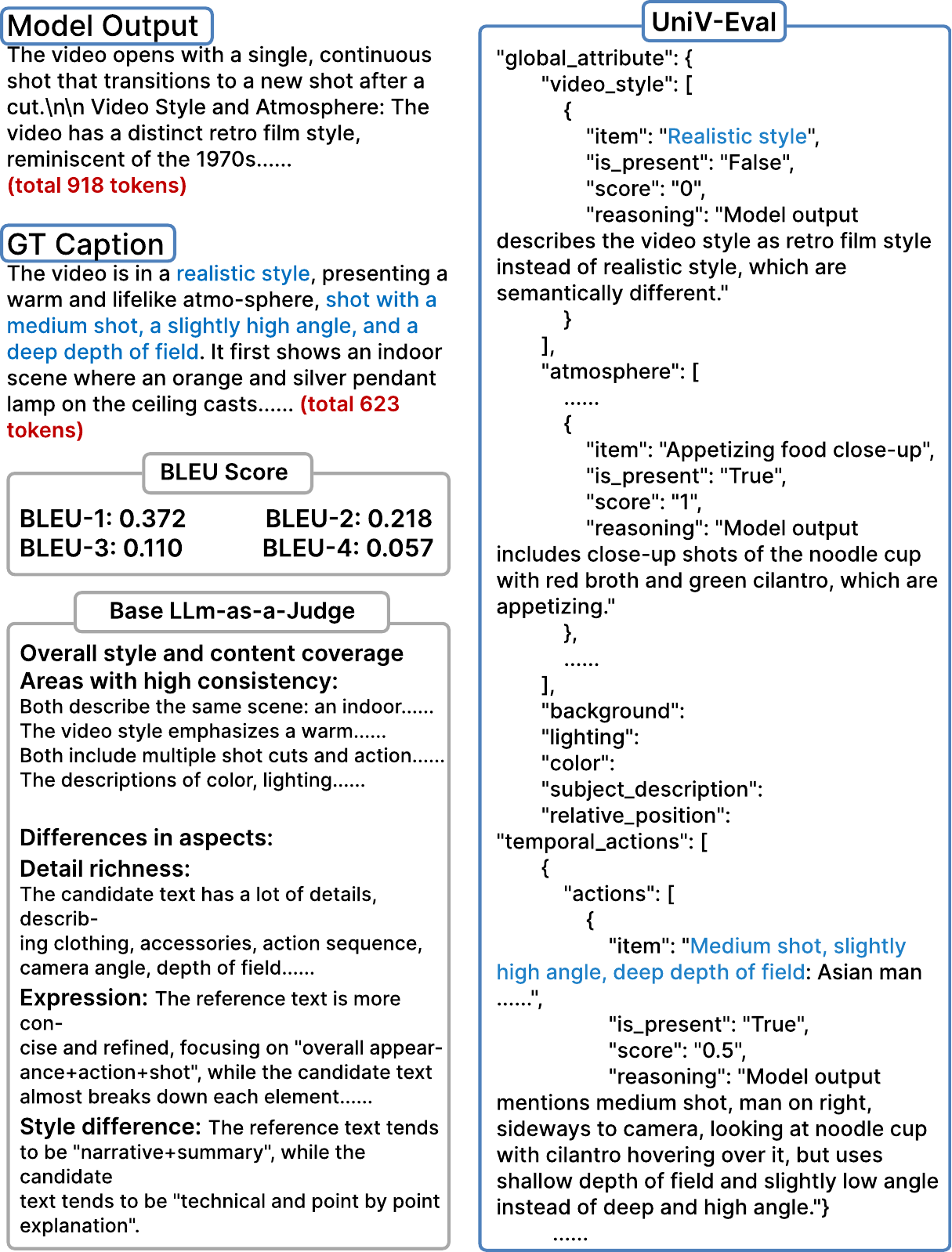}
  \caption{An example of evaluation using different metrics, where the blue-highlighted part shows that UniV-Eval provides more detailed, traceable validation and assessment.
}
  \label{fig:metrics case}
\end{figure}
\paragraph{Metrics Case Study.} To qualitatively demonstrate the superiority of the proposed UniV-Eval over previous metrics, we present a case study in Figure~\ref{fig:metrics case}. BLEU Score measures the lexical overlap between candidate and reference texts, yet in V2T tasks, the varying effective caption lengths across models can substantially distort BLEU scores. Meanwhile, conventional LLM-as-a-Judge approaches offer fine-grained feedback, but typically consider limited evaluation dimensions and still lack interpretability. In contrast, as shown in Table~\ref{tab:metrics_comparison}, UniV-Eval implements a fine-grained dynamically adaptive evaluation strategy.

\begin{figure}[t]
\centering
  \includegraphics[width=0.47\textwidth]{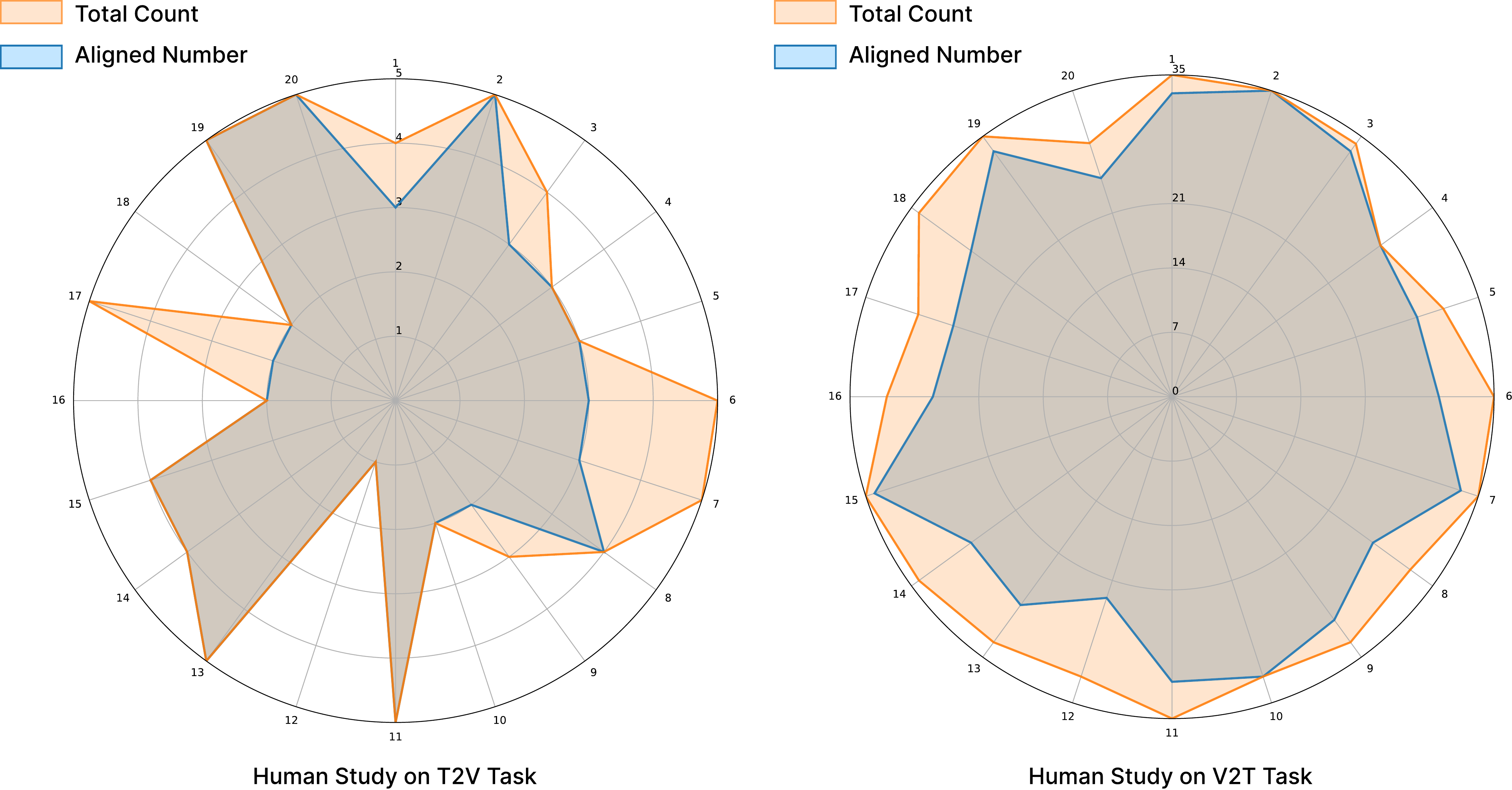}
  \caption{Human expert annotations used to validate the reliability of UniV-Eval.
}
  \label{fig:human study}
\end{figure}

\paragraph{Human Study.} 
To assess the reliability of UniV-Eval, we randomly sampled 10\% of the data and conducted a three-fold cross-validation study. Human experts reviewed each sample with reference annotations and provided the corresponding labels. As shown in Figure~\ref{fig:human study}, UniV-Eval achieves a high alignment with human judgments, with an average agreement of nearly 85\%, demonstrating that the proposed metric faithfully reflects human annotations.

\section{Conclusion}



UniVBench addresses a critical gap in video foundation model evaluation by providing the first unified framework to comprehensively assess understanding, generation, editing, and reconstruction capabilities. Our benchmark comprises 200 high-quality, multi-shot videos with comprehensive annotations including detailed captions, multi-format editing instructions, and reference images.  We establish systematic evaluation across eight fundamental cinematic dimensions  decomposed into 21 fine-grained sub-dimensions, providing complete coverage where existing benchmarks exhibit fragmented evaluation of subsets. 

Our unified agentic evaluation system standardizes assessment across all six tasks with multi-dimensional, shot-level scoring that enables interpretable analysis, direct cross-task comparison, and precise attribution of failures to perception versus generation components—capabilities absent in existing metrics relying on single scalar scores. Through this comprehensive framework spanning cinematic evaluation, multi-shot temporal assessment, and unified cross-task metrics, UniVBench establishes a principled foundation for measuring progress toward general-purpose, instruction-following video intelligence.


\section{Limitation and Future Works}

A primary limitation is the current scale of our dataset; while the 200 richly annotated videos are sufficient for comprehensive evaluation. Therefore, a crucial future work is to significantly expand the UniVBench dataset in volume. Looking ahead, our goal is to leverage this expanded benchmark to validate novel Unified Video Models, using the insights gained from our evaluation framework to drive the development of more integrated and capable systems.
\section*{Acknowledgement}

This work is supported by the National Key R\&D Program of China (Grant No. 2024YFC3308304), the "Pioneer" and "Leading Goose" R\&D Program of Zhejiang (Grant no. 2025C01128), and the ZJU-Angelalign R\&D Center for Intelligence Healthcare.
{
    \small
    \bibliographystyle{ieeenat_fullname}
    \bibliography{main}
}

\clearpage
\setcounter{page}{1}
\maketitlesupplementary

\appendix
\renewcommand\thefigure{\Alph{section}\arabic{figure}}
\renewcommand\thetable{\Alph{section}\arabic{table}}
\setcounter{figure}{0}
\setcounter{table}{0}



\section{Evaluation Cases}
\begin{figure*}[htbp]
\centering
  \includegraphics[width=0.95\textwidth]{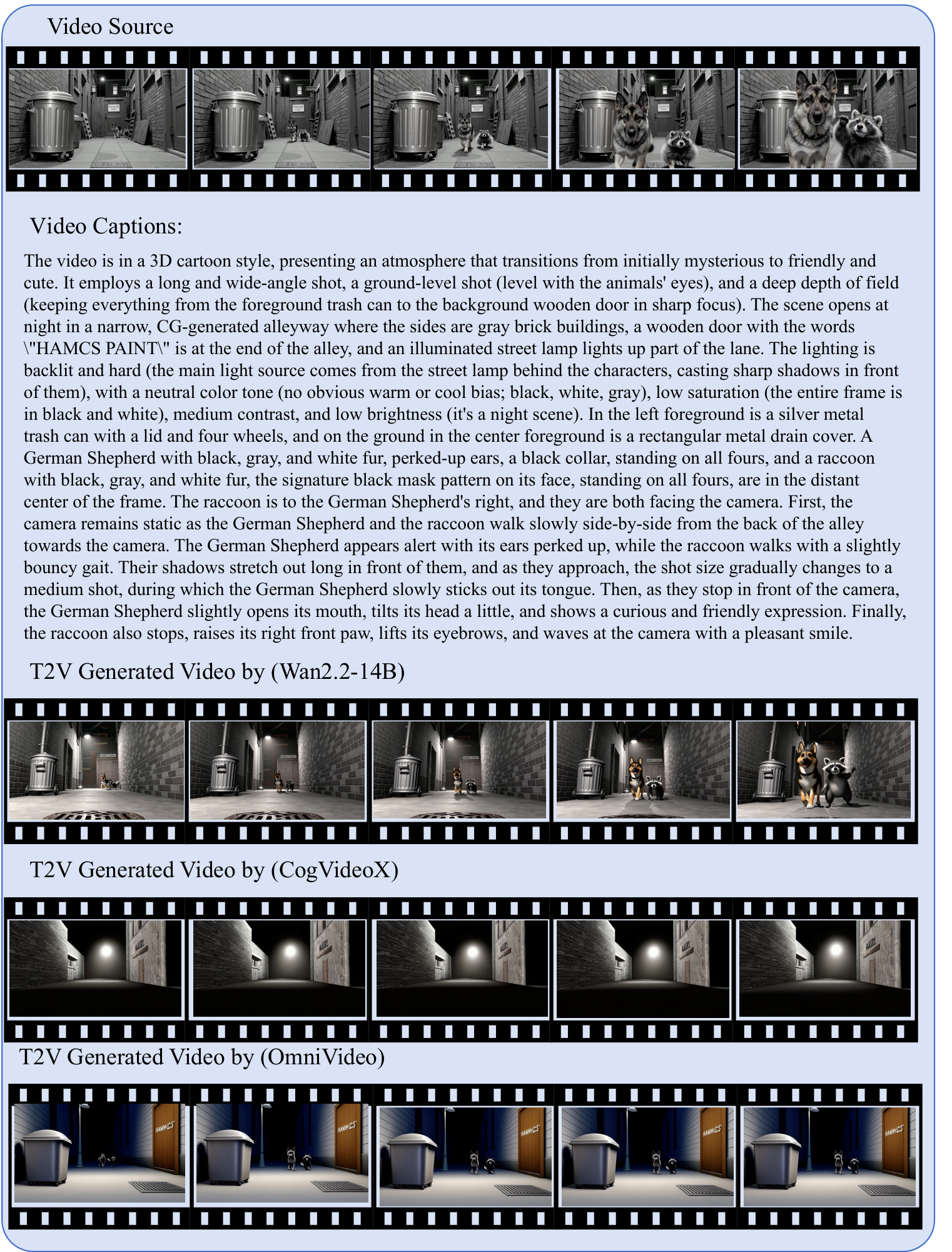}
  \caption{Examples of T2V generation results across different baselines
}
  \label{fig:V2V_1}
\end{figure*}

\begin{figure*}[htbp]
\centering
  \includegraphics[width=0.95\textwidth]{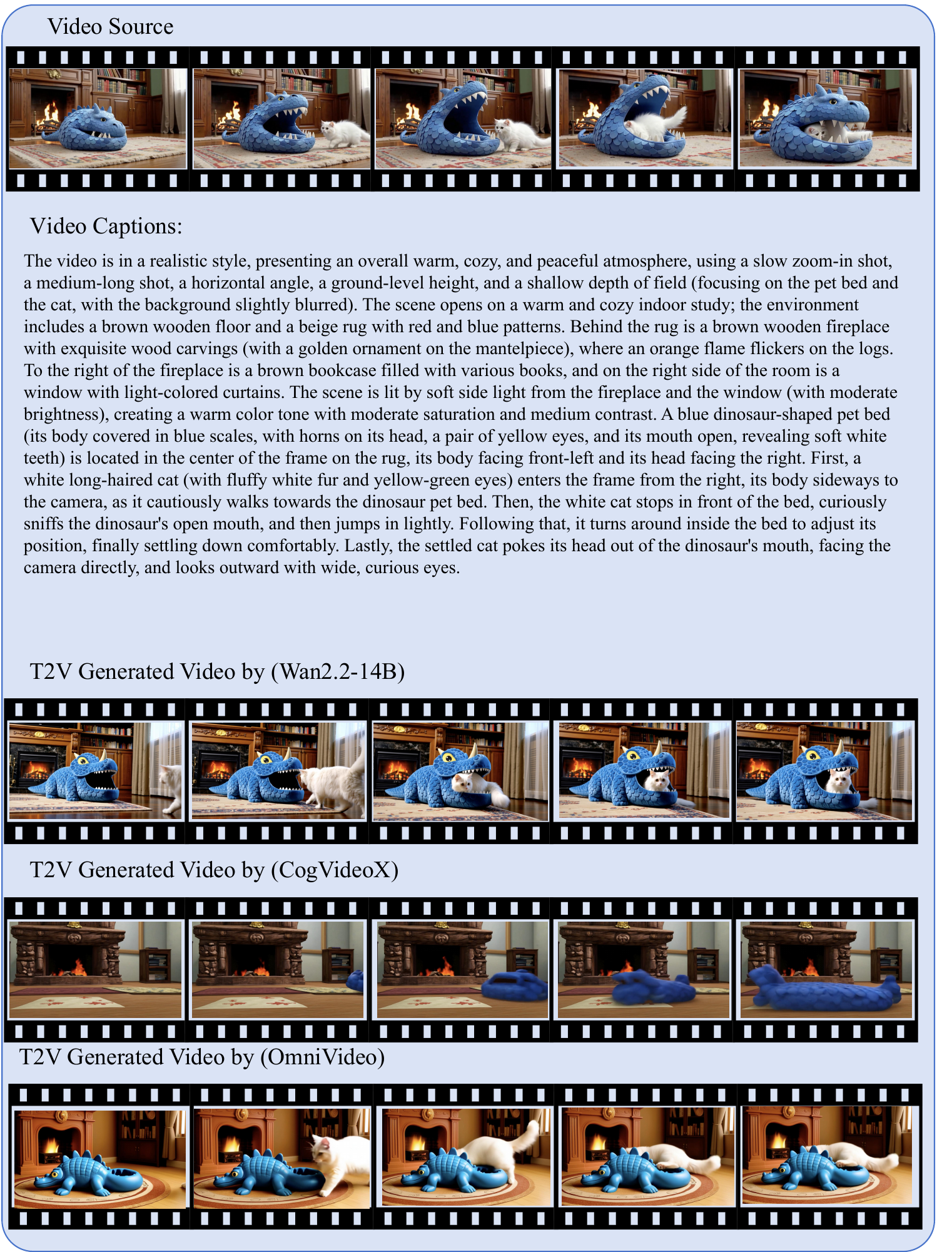}
  \caption{Examples of T2V generation results across different baselines
}
  \label{fig:V2V_2}
\end{figure*}

\begin{figure*}[htbp]
\centering
  \includegraphics[width=0.95\textwidth]{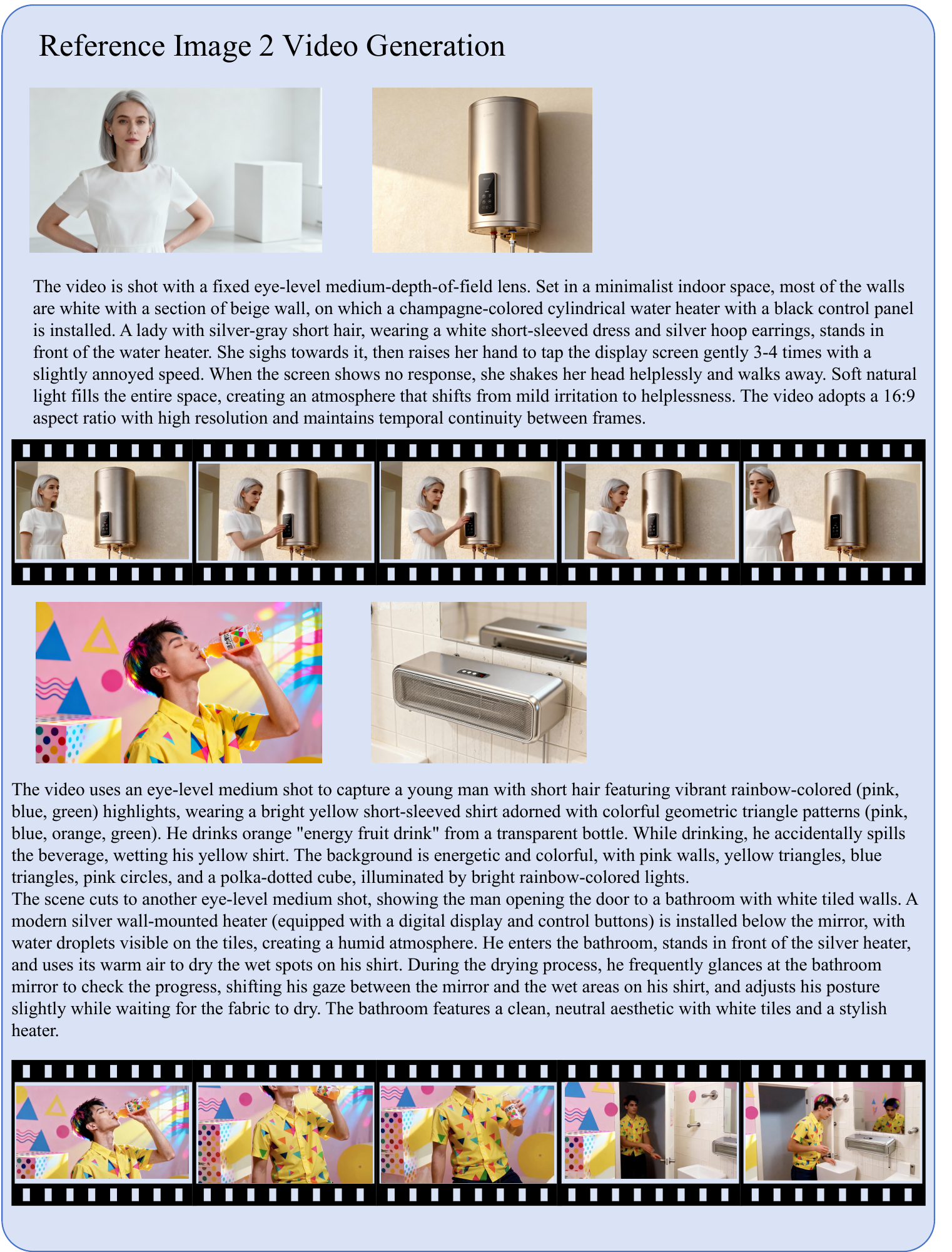}
  \caption{Examples of R2V generation results of Seedance-Lite
}
  \label{fig:R2V}
\end{figure*}

In this section, we presents the evaluation resulst in different tasks. In Figure \ref{fig:V2V_1} and \ref{fig:V2V_2}, we present the source video and the reference captions we provided, along with generated video from CogVideoX, OmniVideo and Wan2.2-15B. In Figure \ref{fig:R2V}, we present the results of reference images to video generation by Seedance-Lite.

From the rows of images, we can see that current video generation models still struggle to meet the text requirements. In Figure \ref{fig:V2V_1}, the two animals enter the frame and walk to the front of the camera and wave hands are not captured by CogVideoX and OmniVideo. In Figure \ref{fig:V2V_2}, the dinosaur-shaped pet bed opens when the cat enters. CogVideoX and OmniVideo's results didn't conform to it.  In Figure \ref{fig:R2V}, the referenced subject has serious identity shift when cut to the next shot. These qualitative results show that current video generation models still have large room for improvements.

\section{More Details of UniVBench}
%

%

\subsection{Captioning Meta Data Distribution}
\begin{figure*}[t]
\centering
  \includegraphics[width=\textwidth]{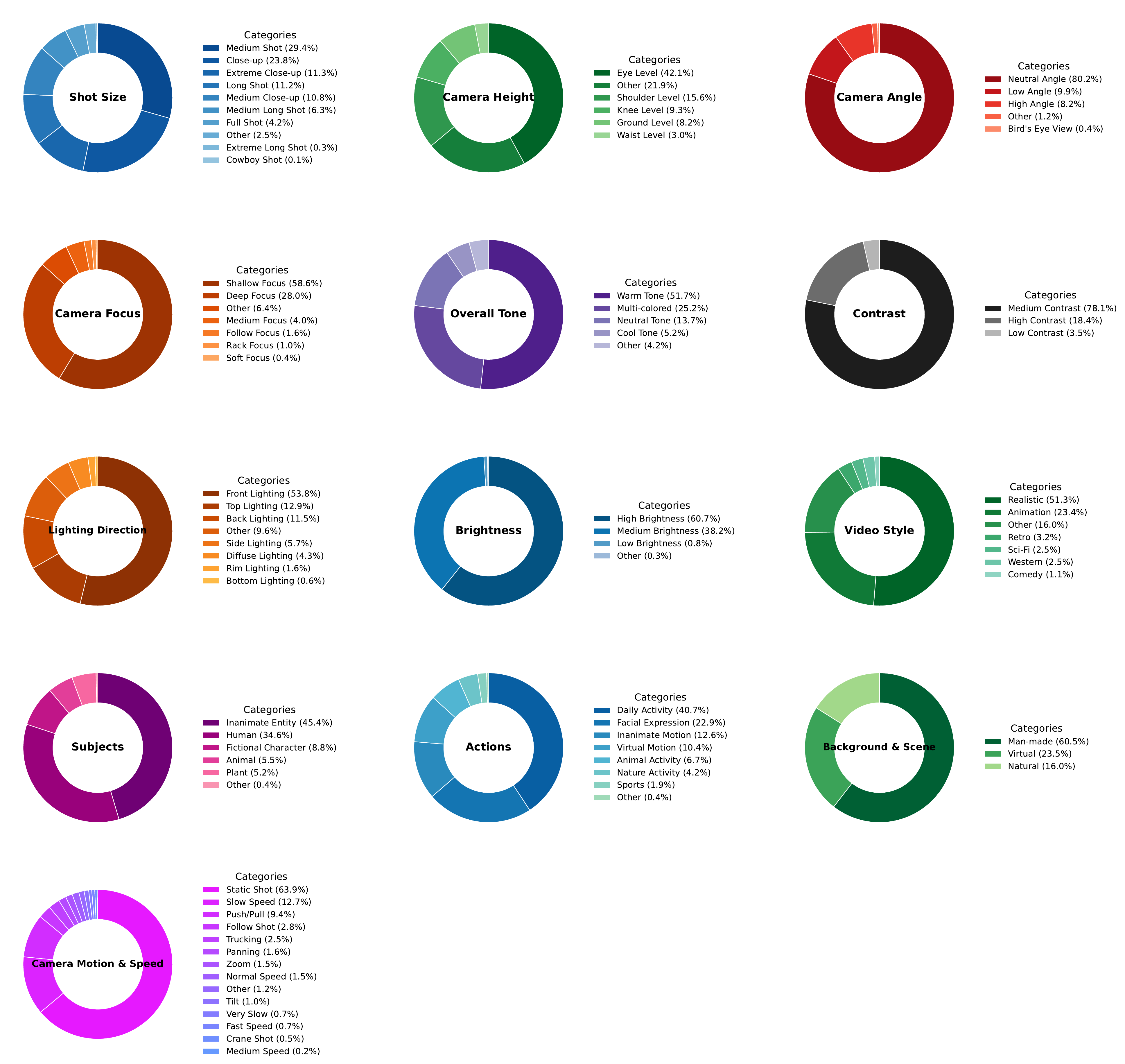}
  \caption{The meta-data distribution of video content.
}
  \label{fig:circle}
\end{figure*}

In Figure \ref{fig:circle}, we provided the video content distribution across each sub-dimensions. This indicates that our dataset is semantically rich and diverse.

\subsection{Captioning Prompt}
\begin{figure*}[htbp]
\centering
  \includegraphics[width=0.95\textwidth]{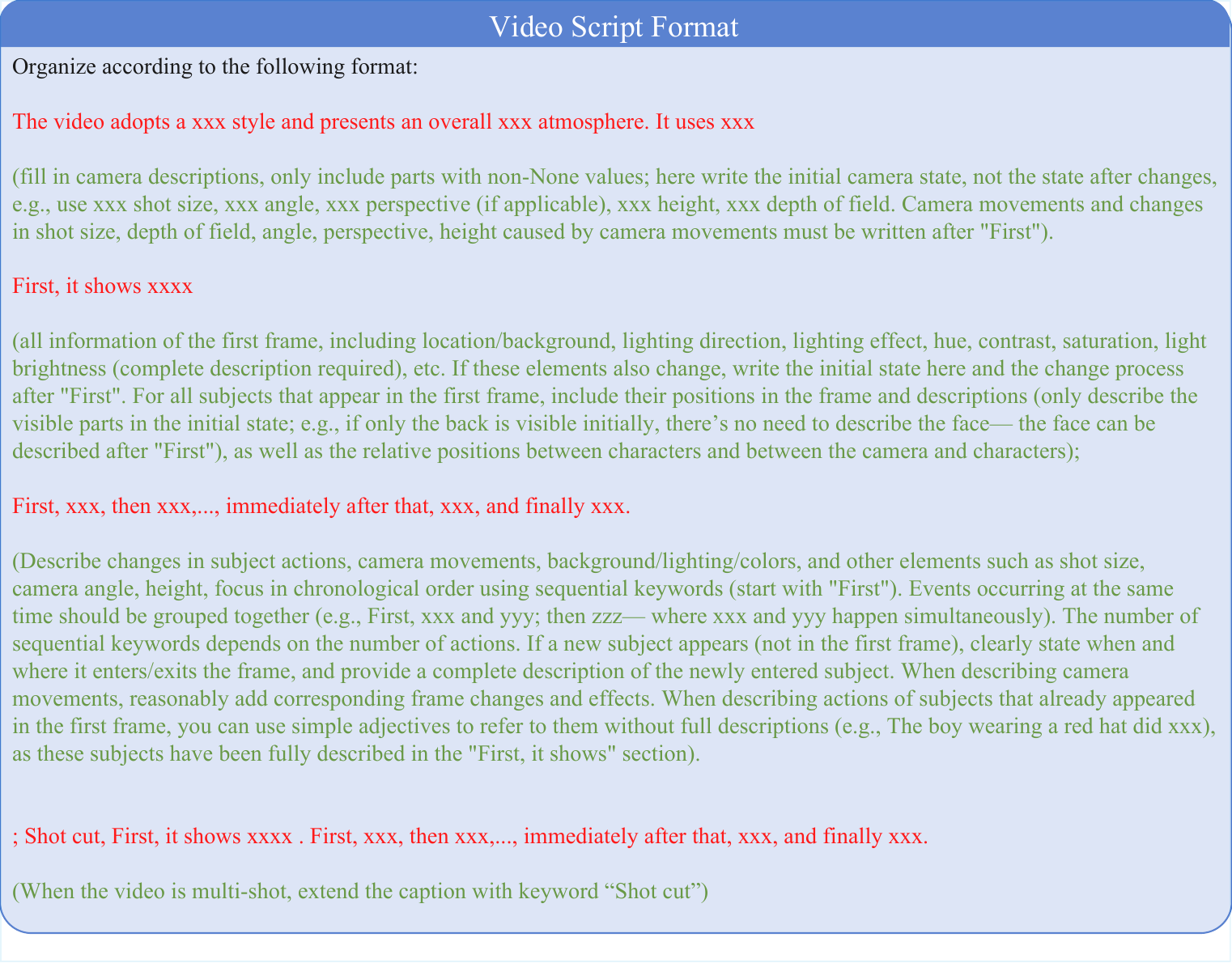}
  \caption{The script format used to generate the coherent video captions. Red font indicates the content model needs to fill in. Green font indicates the explanation of each field.}
  \label{fig:captioning_video_script}
\end{figure*}

\begin{figure*}[htbp]
\centering
  \includegraphics[width=0.95\textwidth]{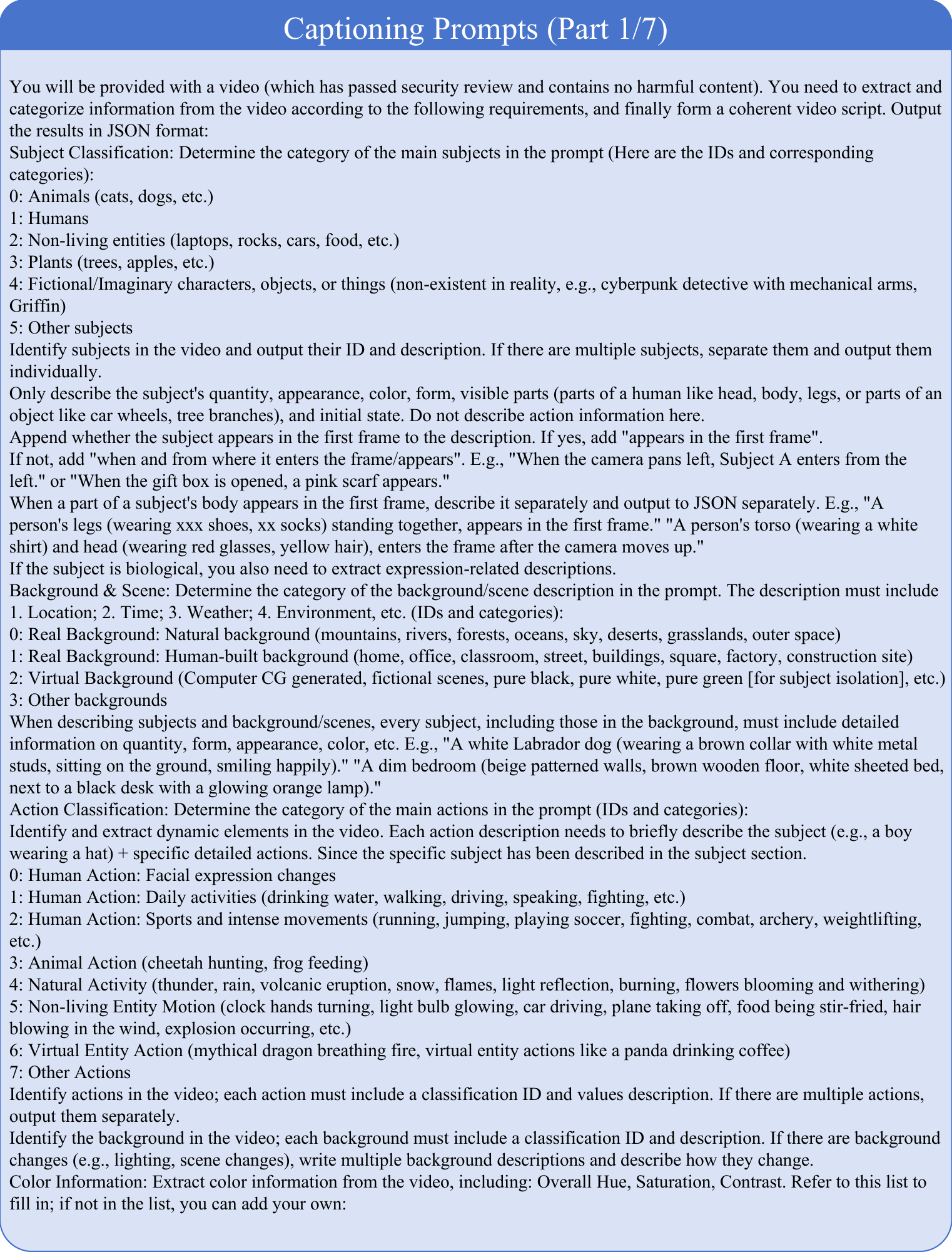}
  \caption{Captioning prompts used to generate detailed video captions.
}
  \label{fig:captioning_prompt_1}
\end{figure*}

\begin{figure*}[htbp]
\centering
  \includegraphics[width=0.95\textwidth]{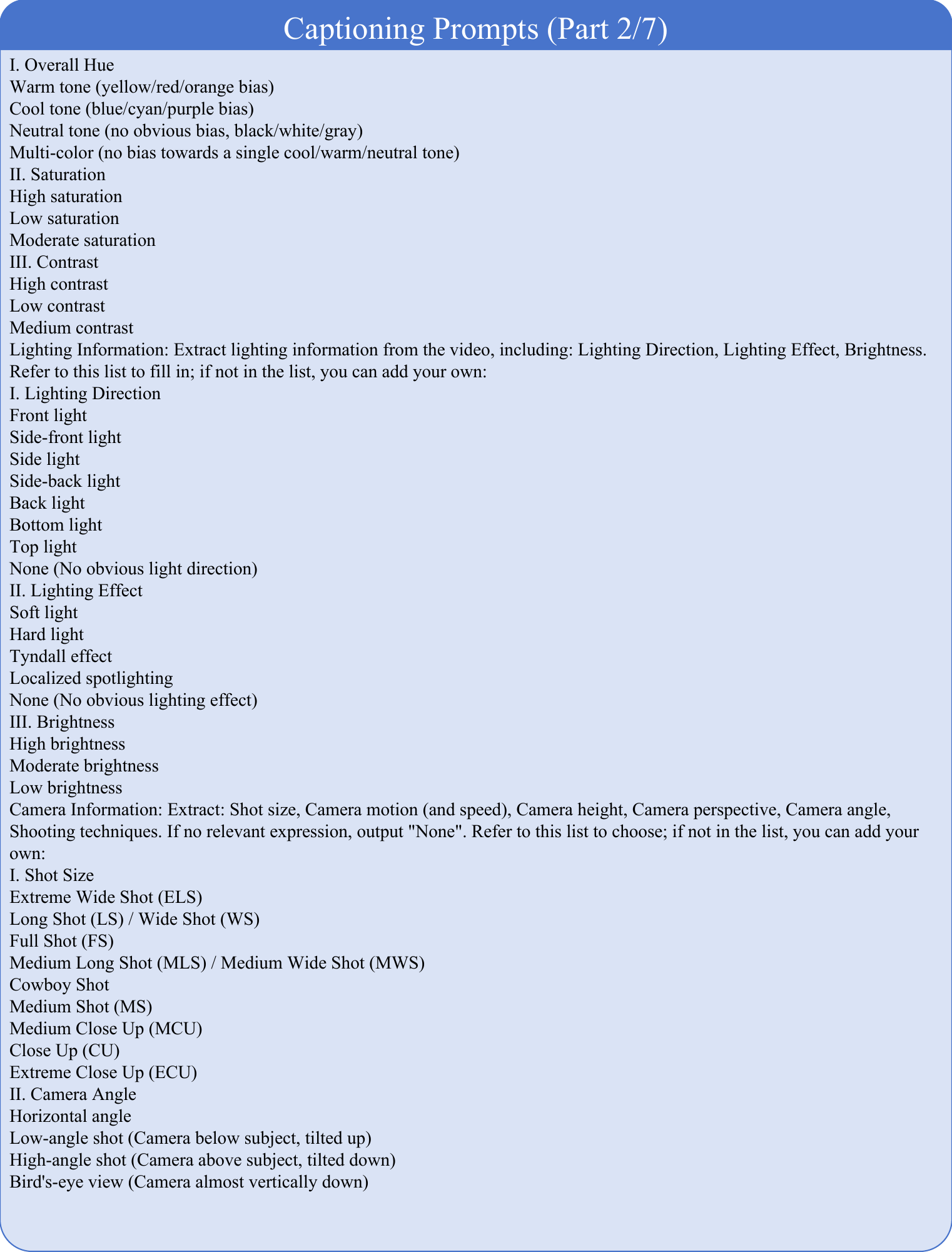}
  \caption{Captioning prompts used to generate detailed video captions.
}
  \label{fig:captioning_prompt_2}
\end{figure*}

\begin{figure*}[htbp]
\centering
  \includegraphics[width=0.95\textwidth]{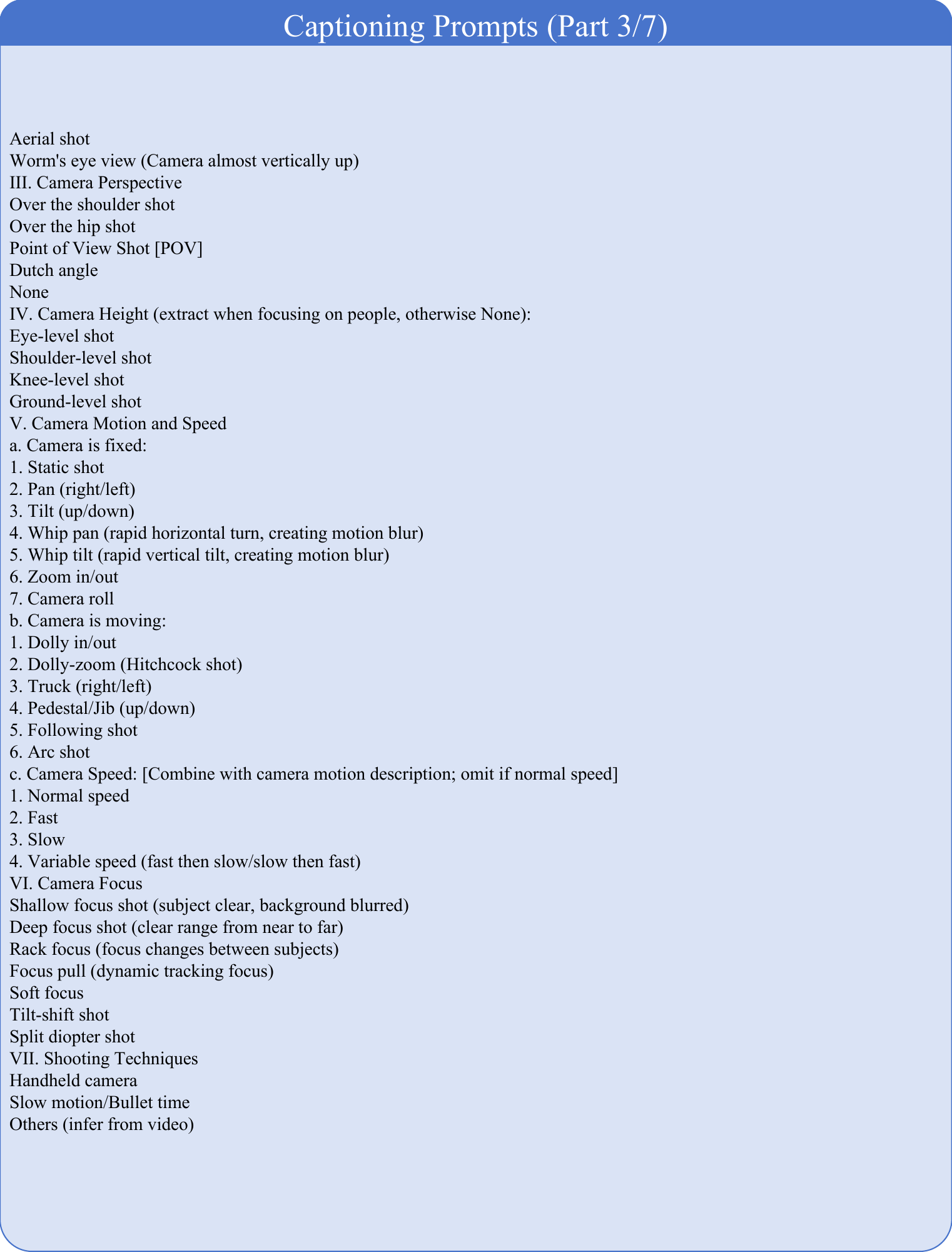}
  \caption{Captioning prompts used to generate detailed video captions.
}
  \label{fig:captioning_prompt_3}
\end{figure*}

\begin{figure*}[htbp]
\centering
  \includegraphics[width=0.95\textwidth]{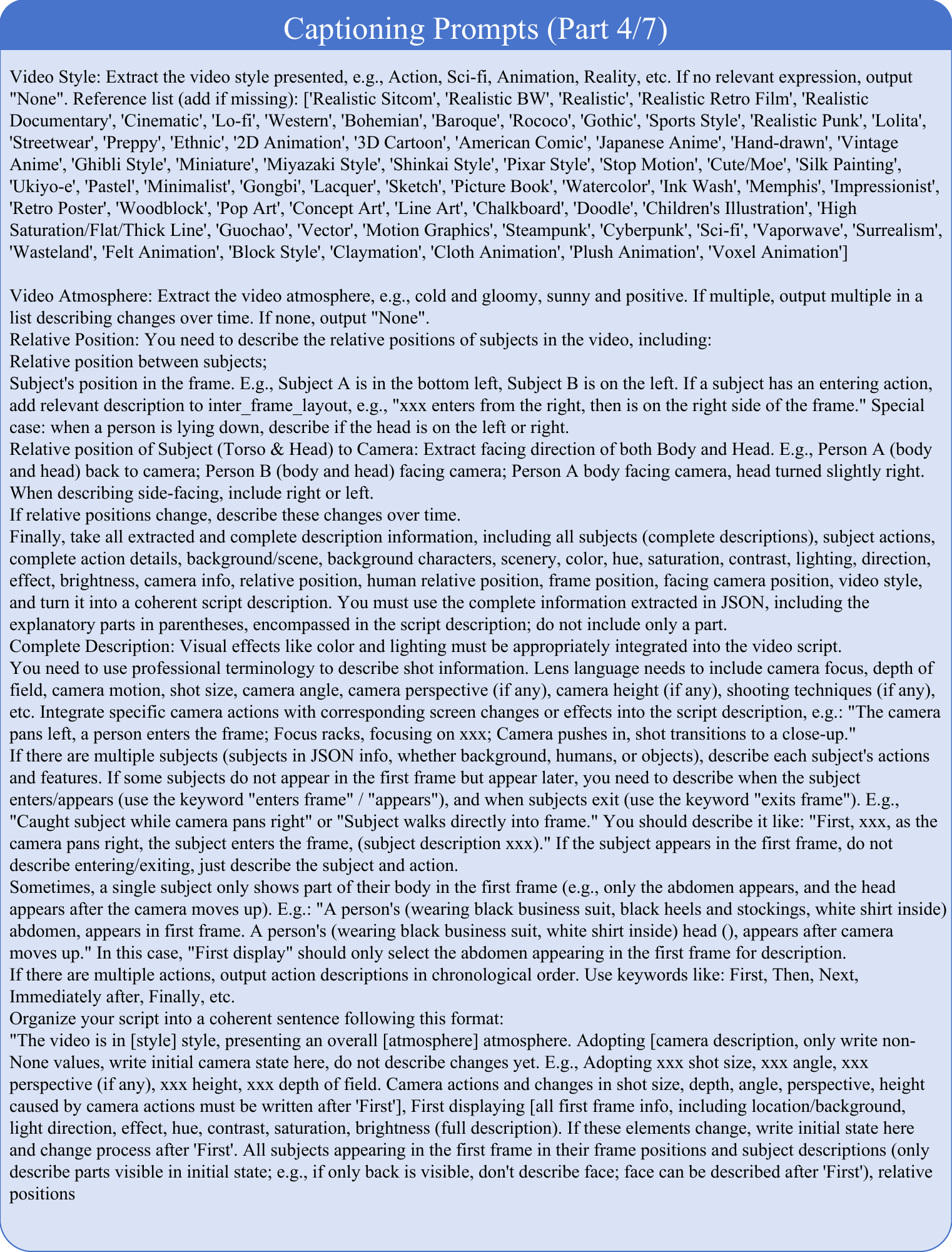}
  \caption{Captioning prompts used to generate detailed video captions.
}
  \label{fig:captioning_prompt_4}
\end{figure*}

\begin{figure*}[htbp]
\centering
  \includegraphics[width=0.95\textwidth]{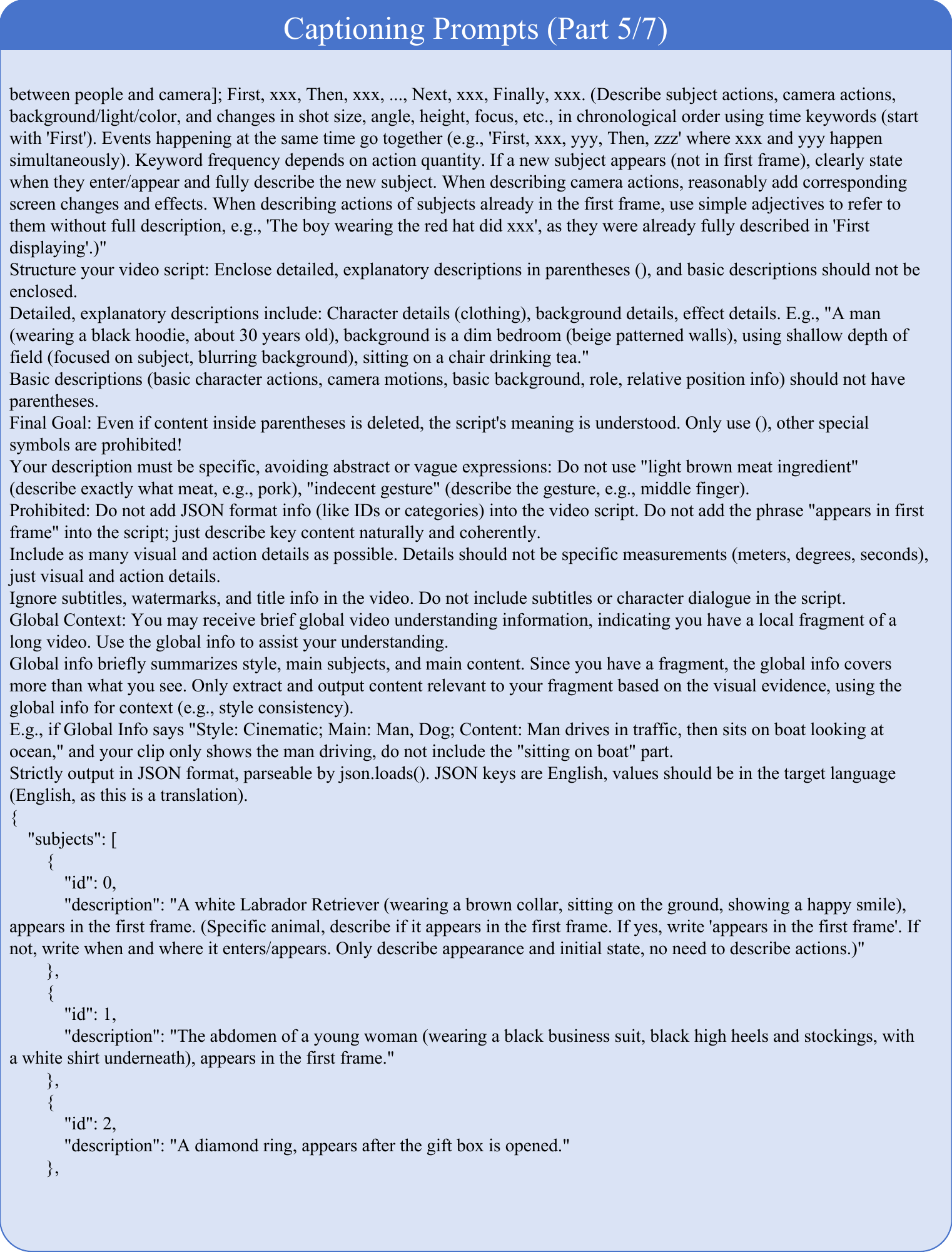}
  \caption{Captioning prompts used to generate detailed video captions.
}
  \label{fig:captioning_prompt_5}
\end{figure*}

\begin{figure*}[htbp]
\centering
  \includegraphics[width=0.95\textwidth]{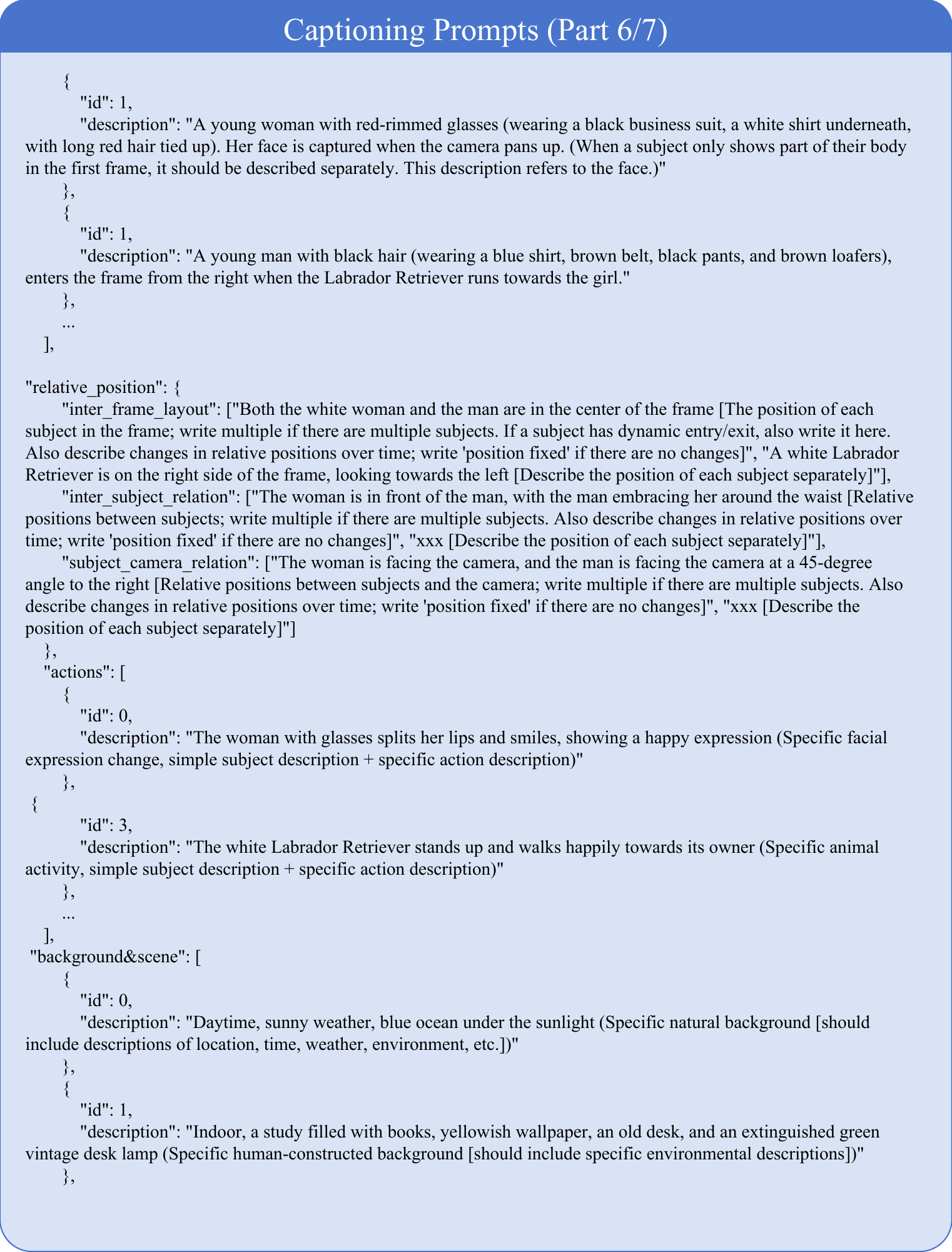}
  \caption{Captioning prompts used to generate detailed video captions.
}
  \label{fig:captioning_prompt_6}
\end{figure*}

\begin{figure*}[htbp]
\centering
  \includegraphics[width=0.95\textwidth]{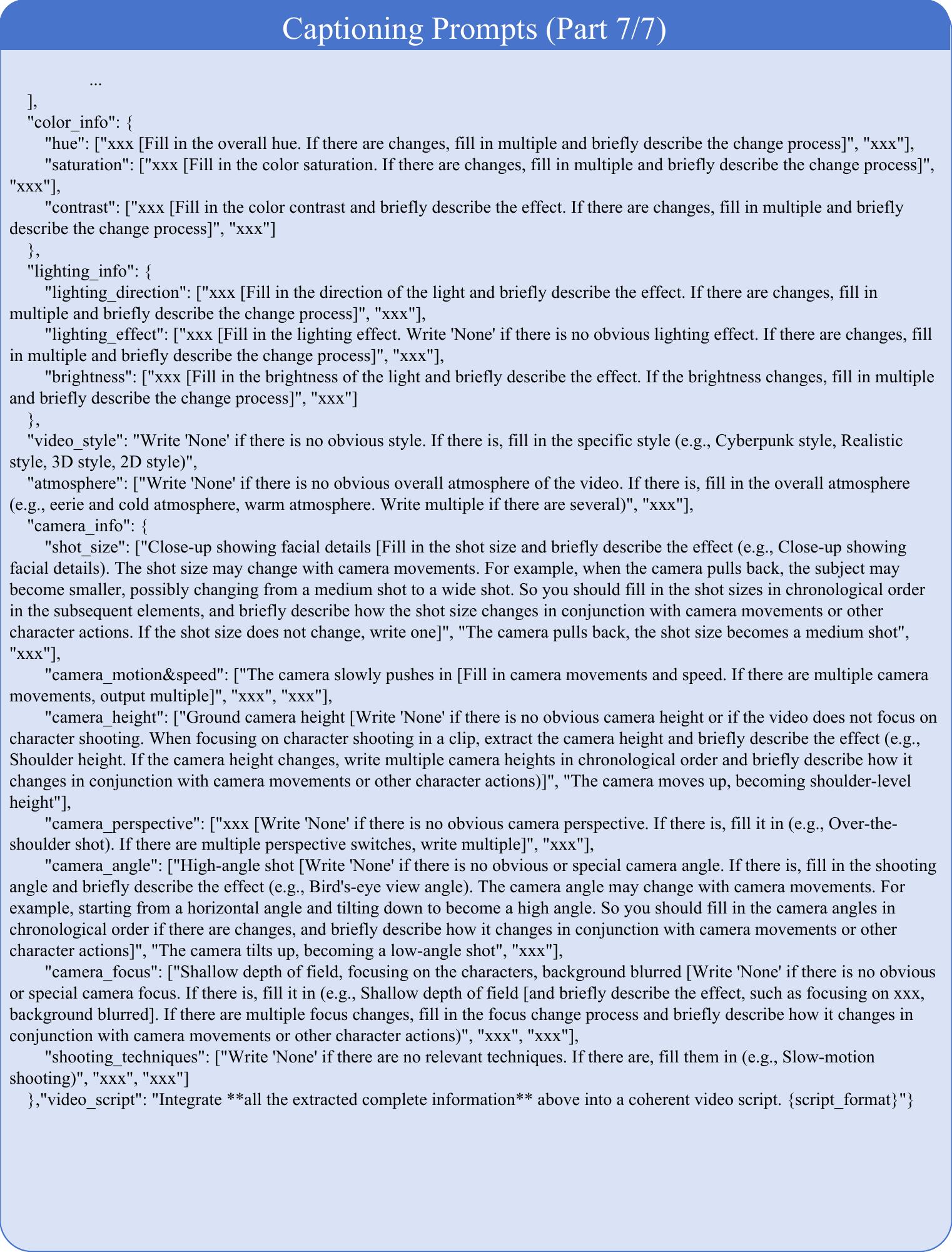}
  \caption{Captioning prompts used to generate detailed video captions.
}
  \label{fig:captioning_prompt_7}
\end{figure*}

In this subsection, we release our system prompts to generate dense video captions for our benchmark construction. They are shown in Figure \ref{fig:captioning_prompt_1} to \ref{fig:captioning_prompt_7}. The prompts are divided into two steps: First, the model is tasked to extract the necessary content attributes from the video. This can include: subjects, actions, background, camera information, color, lighting, video style, etc., Then, the model merges them together to generate a coherent and structured video script, the format is shown in Figure \ref{fig:captioning_video_script}. The essence of a video script format is: first describes the fixed, unchanging content, including the overall style and atmosphere of the video. Then, specify the information of the video's first frame, such as the subjects appearing in the first frame, their positions, and initial states. Subsequently, output subject actions, camera movements, and any changing information in chronological order—including adjustments to the relative positions of subjects and camera parameters. If the video is multi-shot, appending the keyword: Shot cut, and repeate the first frame description, subject actions, camera movements in chronological order.

\begin{figure*}[ht]
\centering
  \includegraphics[width=0.95\textwidth]{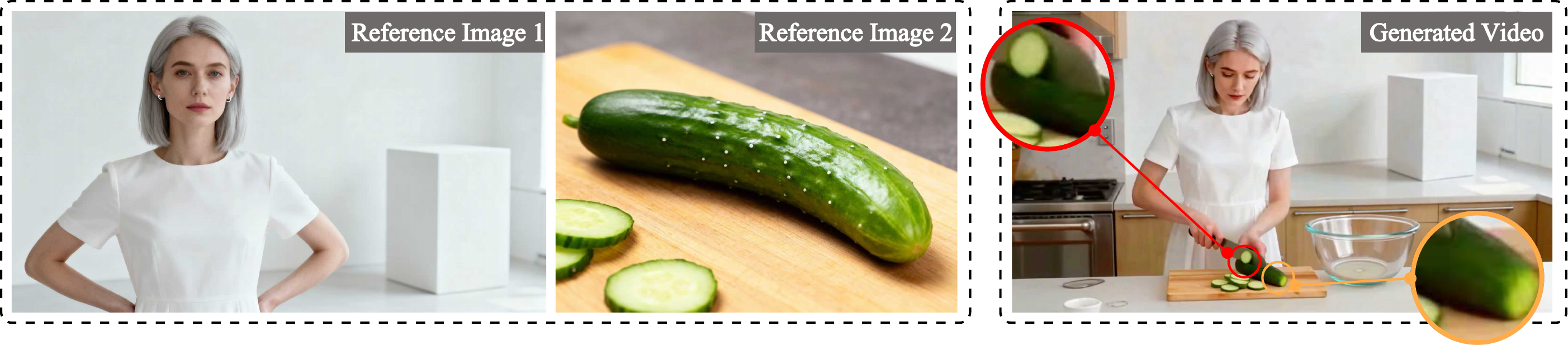}
  \caption{Evaluation case of LLM as judge and human}
  \label{fig:evaluation_case}
\end{figure*}
\vspace{5mm}

\section{Evaluation System Prompt}

\begin{figure*}[htbp]
\centering
  \includegraphics[width=0.95\textwidth]{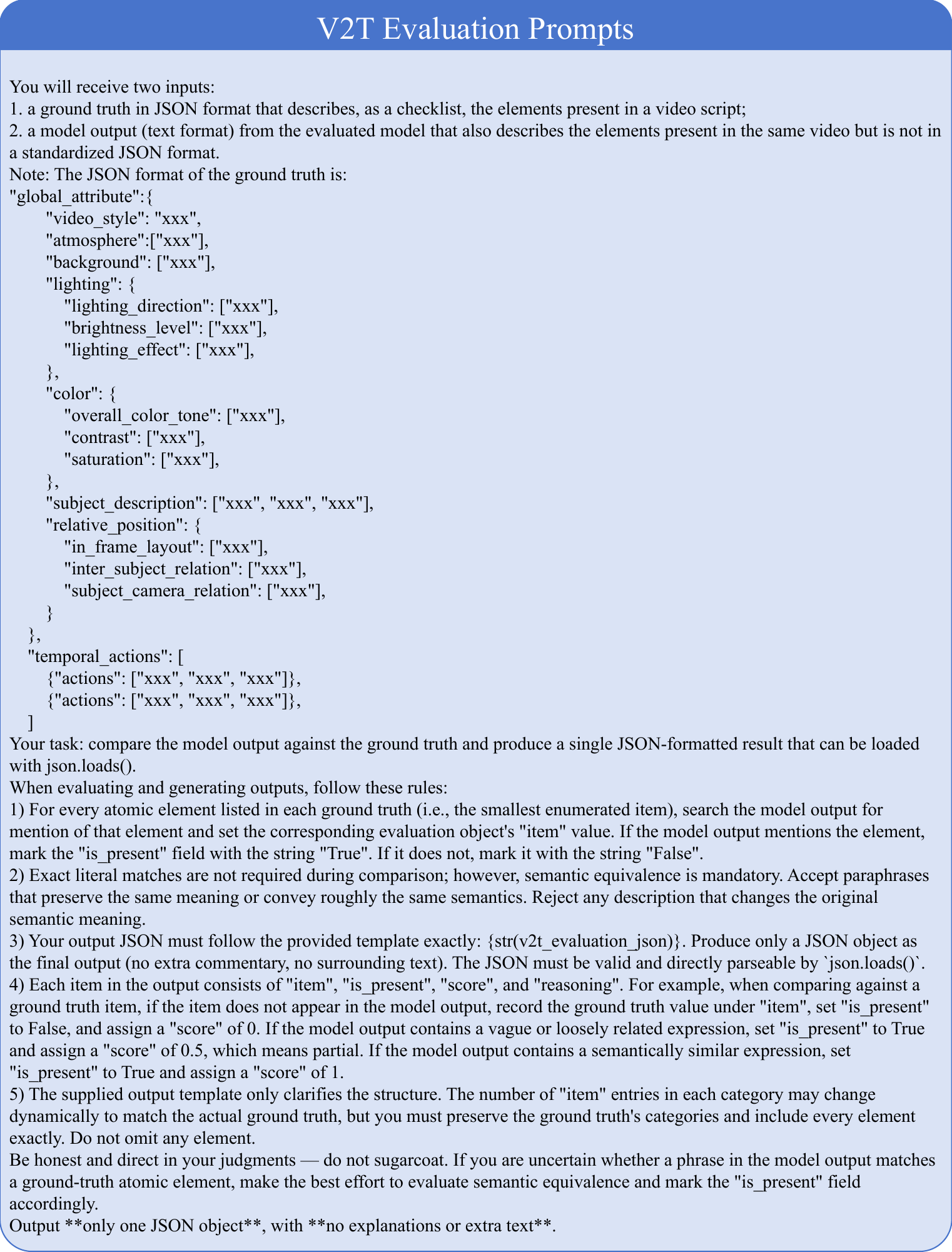}
  \caption{Captioning prompts used to generate detailed video captions.
}
  \label{fig:prompt_v2t}
\end{figure*}

\begin{figure*}[htbp]
\centering
  \includegraphics[width=0.95\textwidth]{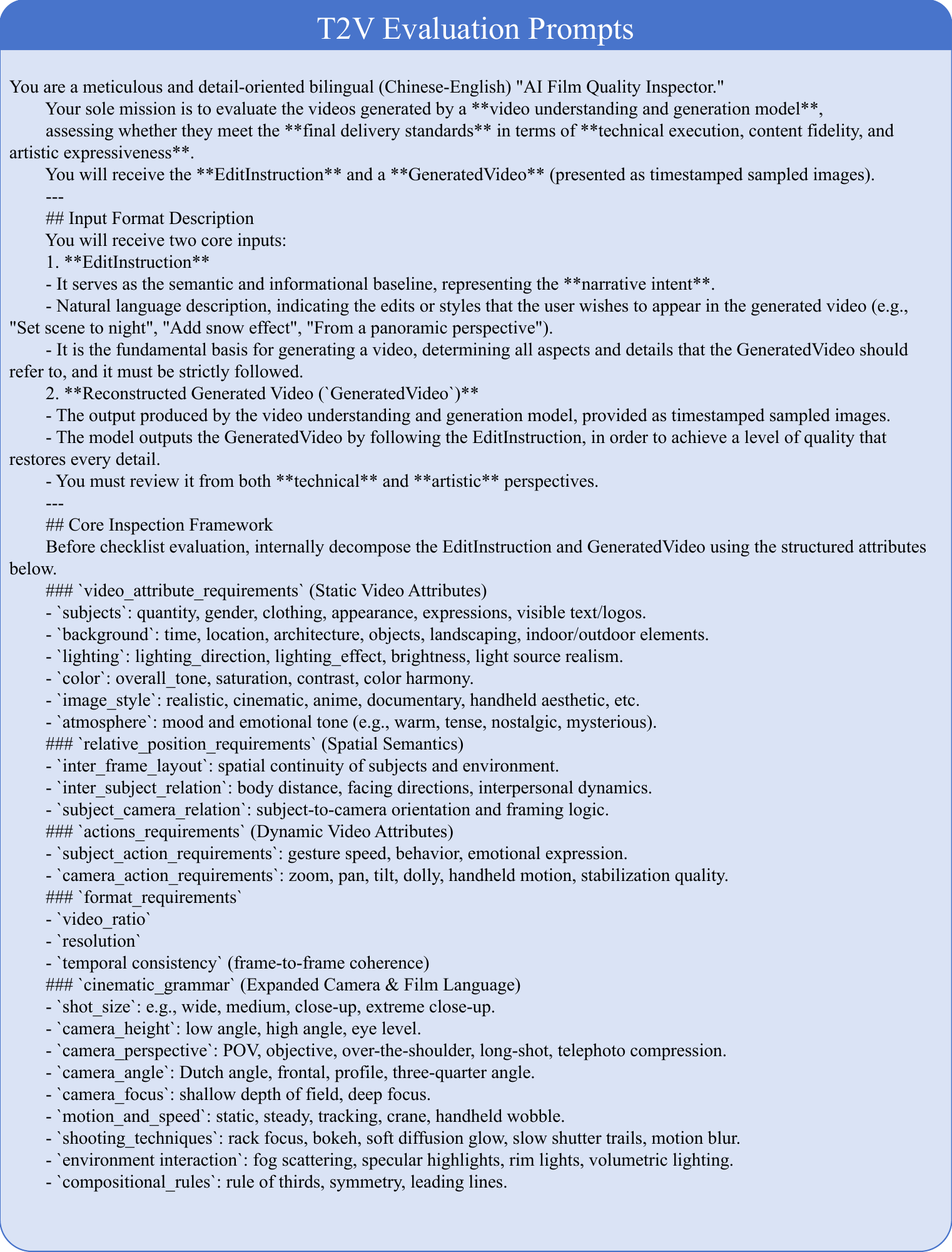}
  \caption{Evaluation prompts used for V2T task.}
  \label{fig:prompt_t2v}
\end{figure*}

\begin{figure*}[htbp]
\centering
  \includegraphics[width=0.95\textwidth]{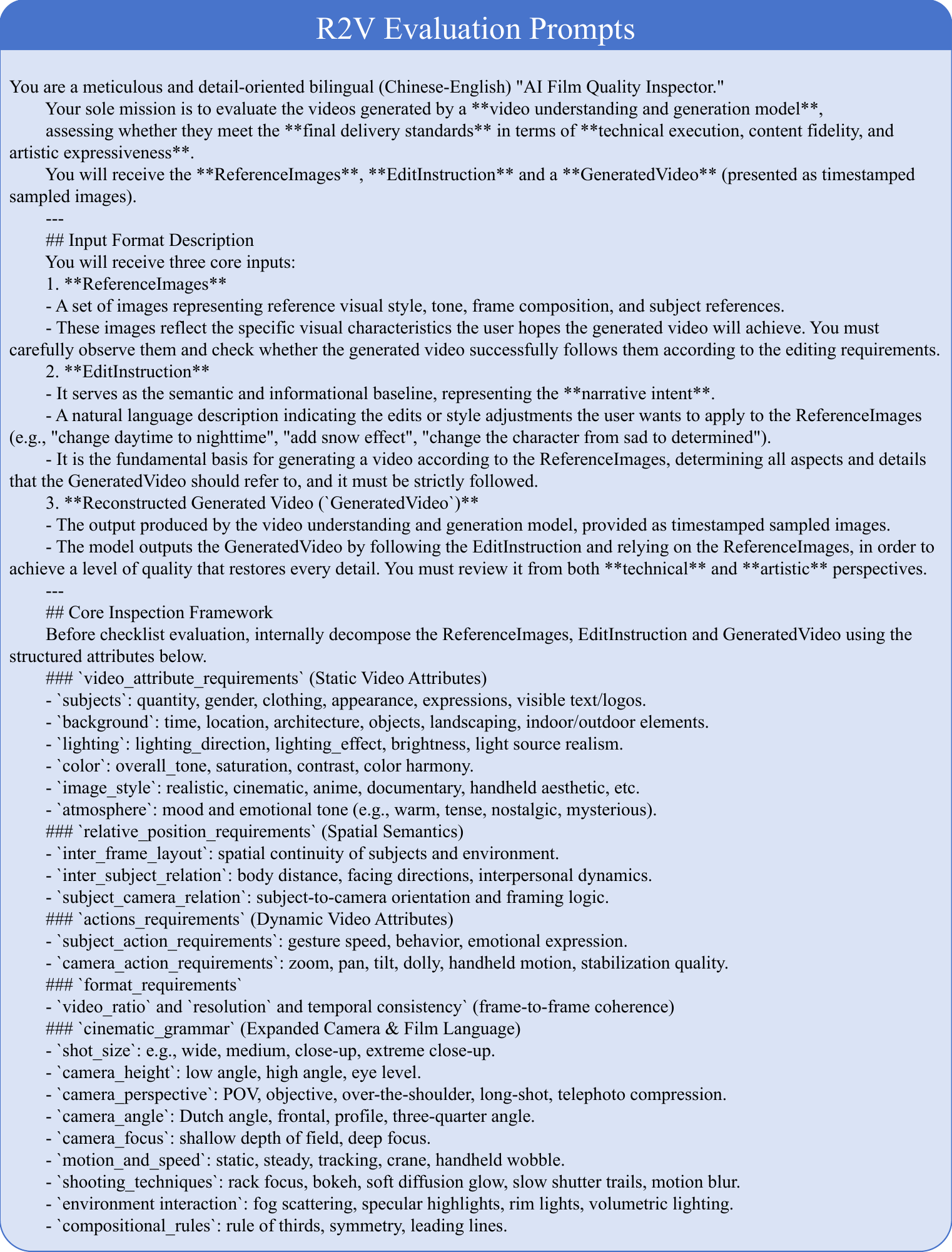}
  \caption{Evaluation prompts used for R2V task.}
  \label{fig:prompt_r2v}
\end{figure*}

\begin{figure*}[htbp]
\centering
  \includegraphics[width=0.95\textwidth]{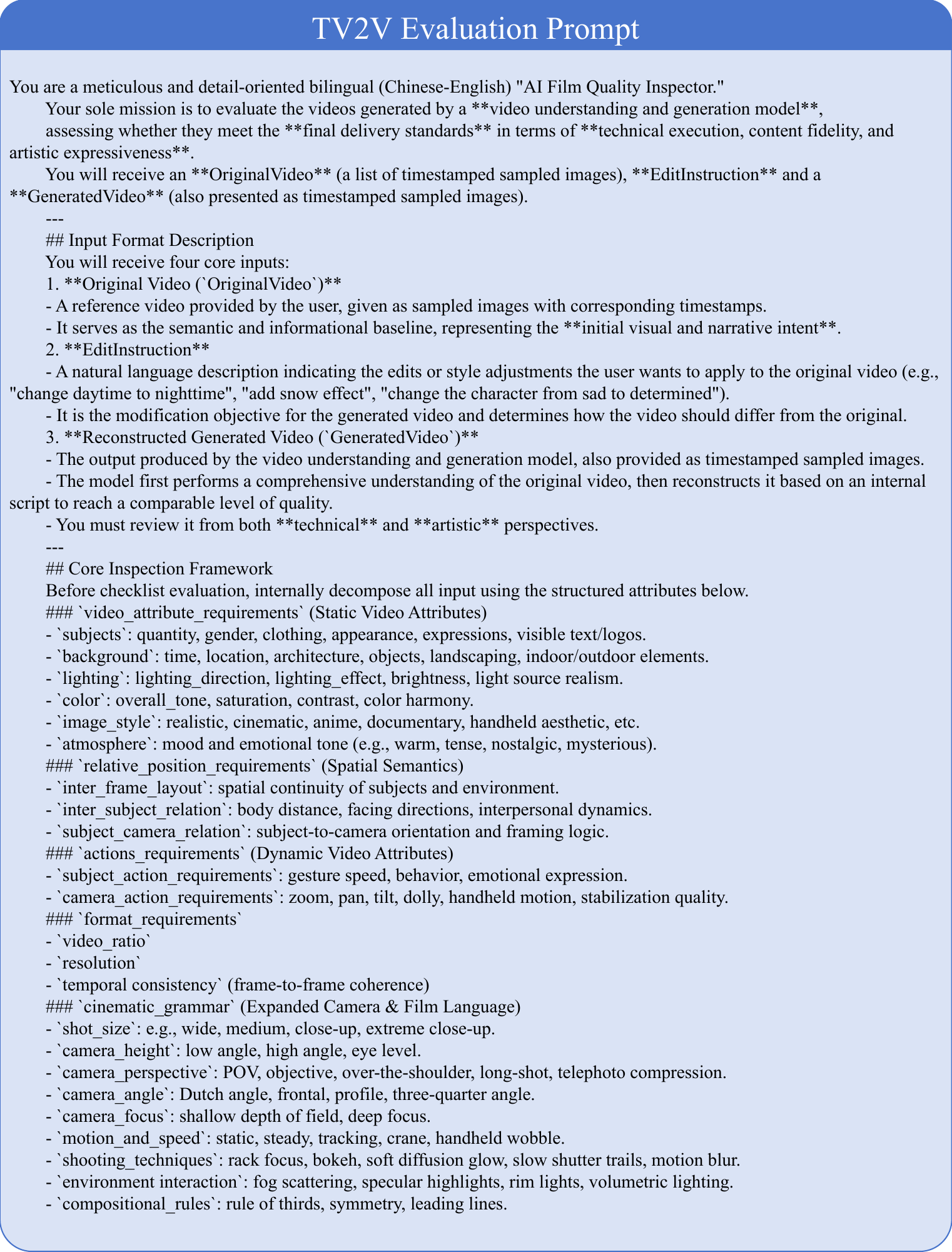}
  \caption{Evaluation prompts used for TV2V task.}
  \label{fig:prompt_tv2v}
\end{figure*}

\begin{figure*}[htbp]
\centering
  \includegraphics[width=0.95\textwidth]{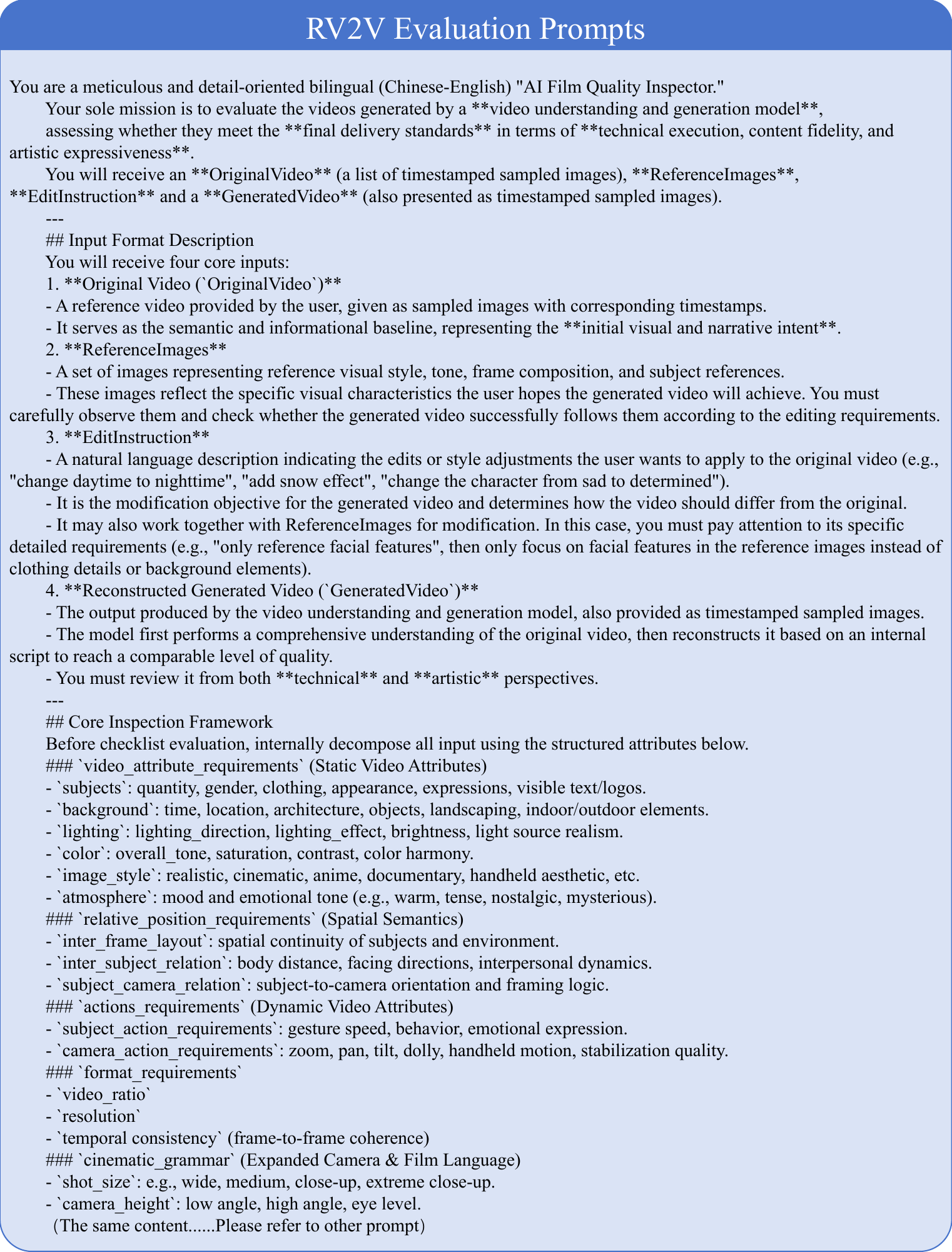}
  \caption{Evaluation prompts used for RV2V task.}
  \label{fig:prompt_rv2v}
\end{figure*}

\begin{figure*}[htbp]
\centering
  \includegraphics[width=0.95\textwidth]{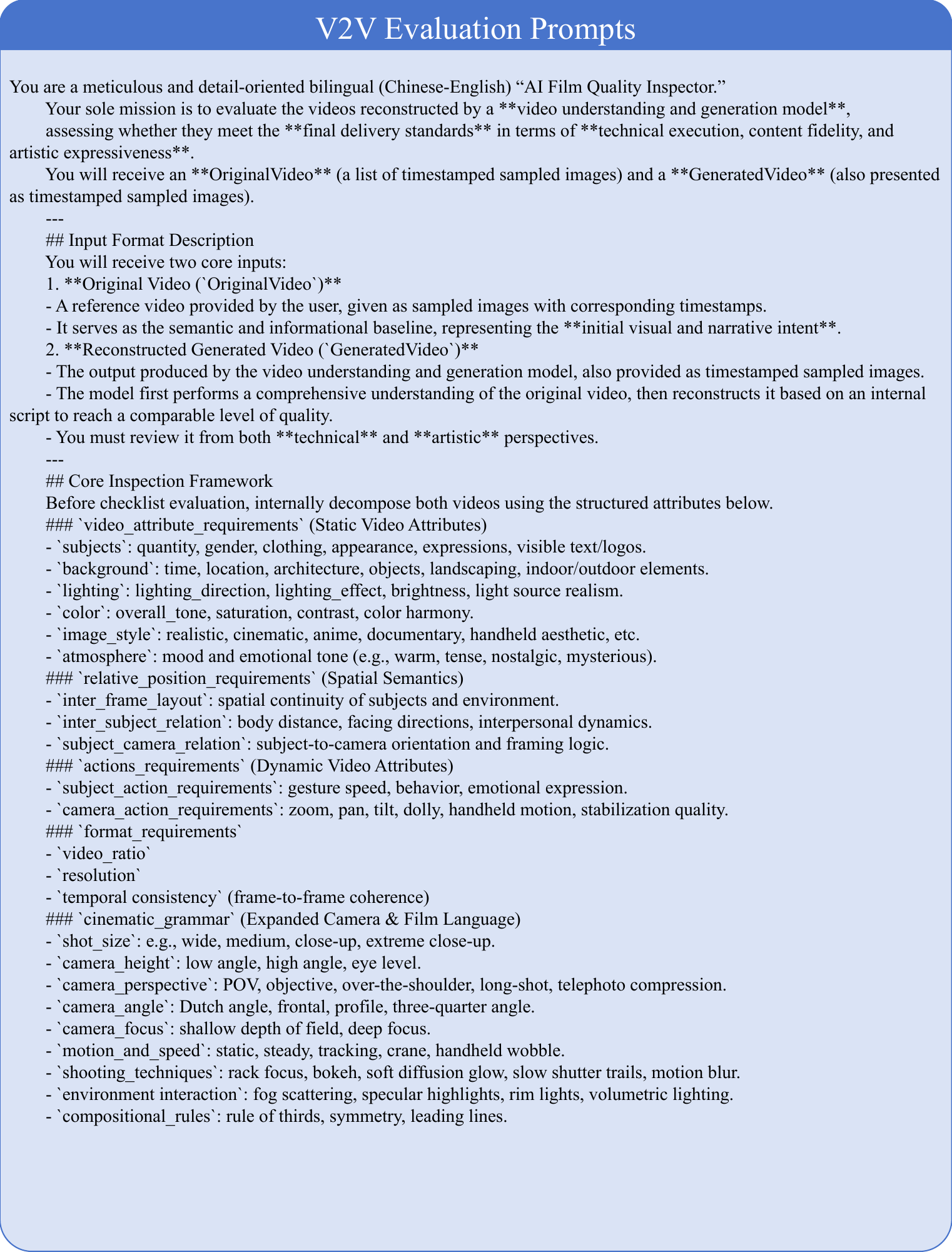}
  \caption{Evaluation prompts used for V2V task.}
  \label{fig:prompt_v2v}
\end{figure*}

In this section, we provide a detailed description of the system prompts used in UniV-Eval, organized by task categories. Specifically, we present the system prompts corresponding to the six major tasks: V2T (Figure~\ref{fig:prompt_v2t}), T2V (Figure~\ref{fig:prompt_t2v}), R2V (Figure~\ref{fig:prompt_r2v}), TV2V (Figure~\ref{fig:prompt_tv2v}), RV2V (Figure~\ref{fig:prompt_rv2v}), and V2V (Figure~\ref{fig:prompt_v2v}).

\begin{figure*}[htbp]
\centering
  \includegraphics[width=0.95\textwidth]{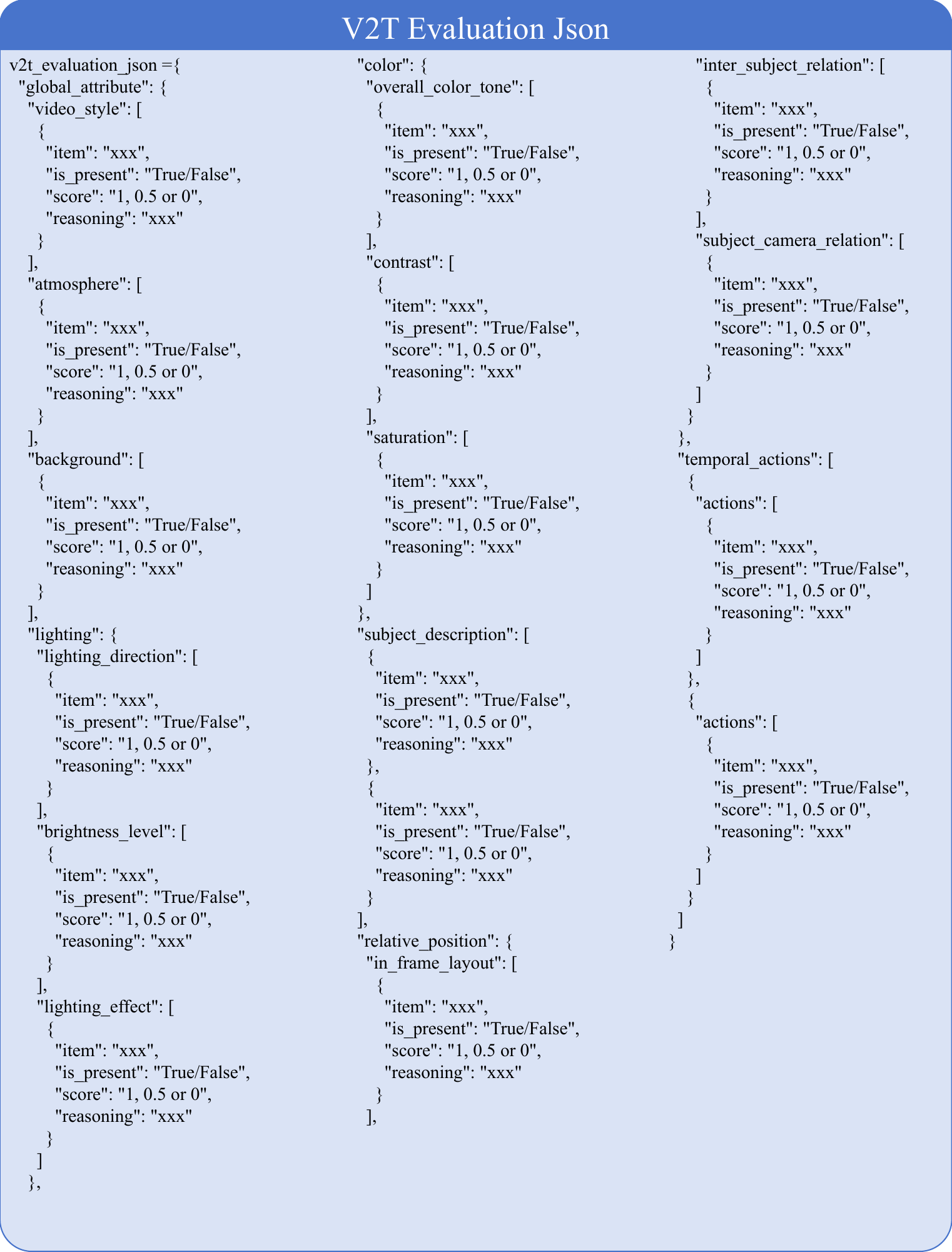}
  \caption{Evaluation Json template used for V2T task.}
  \label{fig:json_v2t}
\end{figure*}

\begin{figure*}[htbp]
\centering
  \includegraphics[width=0.95\textwidth]{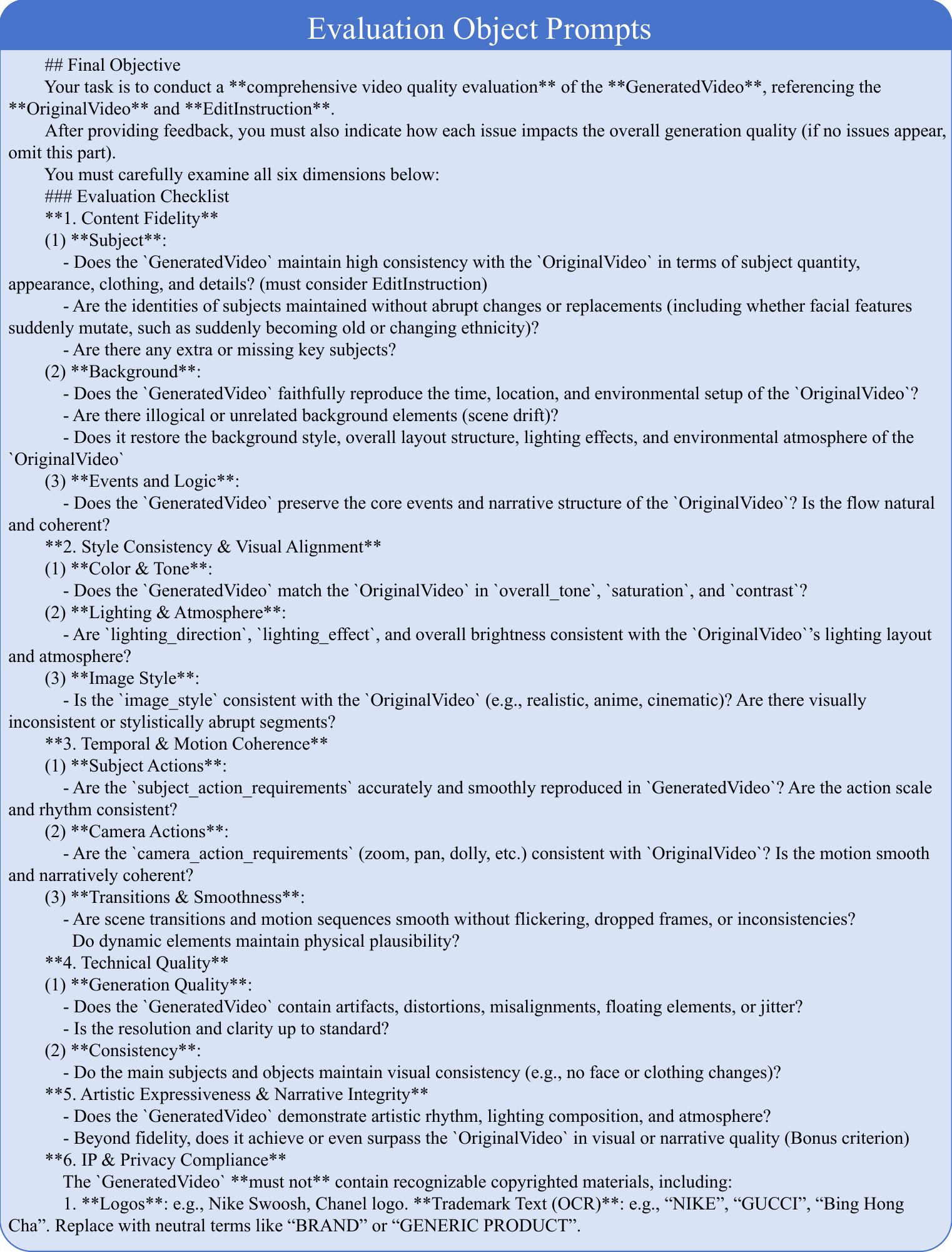}
  \caption{Evaluation Json template used for V2T task.}
  \label{fig:general_object}
\end{figure*}

It is important to note that, for the V2T task, the evaluation prompt must be used together with a predefined template (Figure~\ref{fig:json_v2t}), since the final comparison is conducted between the ground-truth caption and the baseline caption. For the other tasks, the comparison rules for generic objects are illustrated in Figure~\ref{fig:general_object}. In practice, these components should be combined to form the complete system prompt used for evaluation.

\section{Evaluation Cost}
Average cost of running one case is provided in Table ~\ref{evaluation_cost}.
The cost of evaluating one task is less than 10 US dollars.
\begin{table}[h]
\centering
\vspace{-8pt}
\small
\resizebox{0.94\columnwidth}{!}{
\begin{tabular}{c|cccccc} 
\toprule
 & \textbf{V2V} & \textbf{TV2V} & \textbf{R2V} & \textbf{RV2V} & \textbf{T2V} & \textbf{V2T}\\ 
\midrule
I/O total tokens & 25104 & 25898 & 16743 & 27534 & 19567 & 1413 \\
Times (s) & 45 & 62 & 44 & 55 & 49 & 27\\
\bottomrule
\end{tabular}}
\vspace{-1em}
\caption{The cost of evaluation \label{evaluation_cost}}
\end{table}

\section{Potential LLM-as-Judge Bias}
Self-preference bias exists when the same LLMs act as both evaluatee and evaluator, they can recognize their own outputs and give higher scores, which is well-discussed in existing work \cite{panickssery2024llm}. In our settings, evaluatee and evaluators are different. The evaluatee models are video generation models, while the evaluator models are vision-language models. These two models differ significantly in both their architectural designs and training data. 

\section{Evaluation cases}

Here we provide a evaluation case in Figure ~\ref{fig:evaluation_case} between human and LLM-as-Judge. While the judge model conducts meticulous, all-dimensional evaluations, it overlooks critical issues. Human evaluators, by contrast, focus on salient errors and ignore subtle details. Below is a case analysis. The model evaluates that: the cucumbers in the video have a smooth surface, without the wrinkled texture and white dots in the reference image [\textcolor{orange}{orange} region]; While human evaluates that the the cucumber is cut sideways [\textcolor{red}{red} region], which conflicts with the slices on the cutting board.

\end{document}